\documentclass[acmlarge,screen,nonacm]{acmart}

\usepackage{graphicx}
\usepackage{longtable}
\usepackage{multirow,colortbl}
\usepackage{titlesec}
\usepackage{url}
\usepackage{adjustbox}
\usepackage{lscape}

\newcommand{\tmpframe}[1]{\fbox{#1}}
\def\BibTeX{{\rm B\kern-.05em{\sc i\kern-.025em b}\kern-.08emT\kern-.1667em\lower.7ex\hbox{E}\kern-.125emX}}

\begin{document}

%
\title{A Review on Methods and Applications in Multimodal Deep Learning}

%
\author{Jabeen Summaira}
\email{11821129@zju.edu.cn}
\author{Xi Li}
\authornote{Corresponding Author: \textbf{xilizju@zju.edu.cn}}
\email{xilizju@zju.edu.cn}
\affiliation{%
	\institution{College of Computer Science, Zhejiang University, China}
	\streetaddress{P.O. Box W-99}
	\city{Hangzhou}
	\postcode{310027}
}

\author{Amin ~Muhammad Shoib}
\affiliation{%
	\institution{School of Software Engineering, East China Normal University}
	\streetaddress{3663 North Zhongshan Road.}
	\city{Shanghai}
	\country{China}}
\email{52184501030@stu.ecnu.edu.cn}

\author{Jabbar Abdul}
\email{Jabbar@zju.edu.cn}
\affiliation{%
	\institution{College of Computer Science, Zhejiang University, China}
	\streetaddress{P.O. Box W-99}
	\city{Hangzhou}
	\postcode{310027}
}
%
\renewcommand{\shortauthors}{J. Summaira, et al.}

%
\begin{abstract}
Deep Learning has implemented a wide range of applications and has become increasingly popular in recent years. The goal of multimodal deep learning (MMDL) is to create models that can process and link information using various modalities. Despite the extensive development made for unimodal learning, it still cannot cover all the aspects of human learning. Multimodal learning helps to understand and analyze better when various senses are engaged in the processing of information. This paper focuses on multiple types of modalities, i.e., image, video, text, audio, body gestures, facial expressions, and physiological signals. Detailed analysis of the baseline approaches and an in-depth study of recent advancements during the last five years (2017 to 2021) in multimodal deep learning applications has been provided. A fine-grained taxonomy of various multimodal deep learning methods is proposed, elaborating on different applications in more depth. Lastly, main issues are highlighted separately for each domain, along with their possible future research directions.

\end{abstract}

%
%


%

%

%
\maketitle

\section{Introduction:}
Multimodal learning proposes that we are able to remember and understand more when engaging multiple senses during the learning process. MMDL technically contains different aspects and challenges like representation, translation, alignment, fusion, co-learning when learning from two or more modalities \cite{cukurova2020promise,hong2020more}. This information from multiple sources is contextually related and occasionally provides the additional necessary information to one another, revealing features that would not be viewable when working with individual modalities. MMDL models combine heterogeneous data from multiple sources, allowing for more appropriate predictions \cite{sengupta2020review}. 
Extracting and presenting relevant information from multimodal data remains an inspirational motive for MMDL research. Merging various modalities to optimize effectiveness is still an appealing challenge. Furthermore, the accuracy and flexibility of multimodal systems are not optimum due to the insufficiency of labeled data.

The recent advances and trends of MMDL are from Audio-visual speech recognition (AVSR) \cite{yuhas1989integration}, multimedia content indexing and retrieval \cite{atrey2010multimodal,snoek2005multimodal}, understanding human multimodal behaviors during social interaction, multimodal emotion recognition \cite{chen2021heu,boateng2020towards}, image and video captioning \cite{bai2018survey,lei2021video}, Visual Question-Answering (VQA) \cite{long2021improving}, multimedia retrieval \cite{souza2021online} to health analysis \cite{yazdavar2020multimodal} and so on.
In this article, we analyzed the latest MMDL models to propose typical models and techniques for advancing the field forward. Various modalities, i.e., image, video, text, audio, body gestures, facial expressions, and physiological signals, are focused. The main goal of MMDL is to construct a model that can process information from different modalities and relate it. MMDL methods and applications are categorized into multiple groups, and the impact of feature extractor, deep learning architecture, datasets, and evaluation metrics are analyzed for each group. Moreover, key features of each model are also discussed to highlight the main contribution of the work model.

\subsection{Contribution and Relevance to other surveys:}

Recently , numerous surveys are already published relating to the topic of multimodal learning. A summarized list of these review articles is presented and analyzed in Table \ref{tab:relatedSurveys}. A summary of modalities used and applications discussed is shown in the table. Our contributions to this article are described in section \ref{contribution}. 
Most of the baseline surveys only focus on models using Image, Video, Text, and Audio modalities, while our survey paper is focused on some additional models and groups of applications using body gestures, facial expressions, and physiological signals. Comparative analysis on experimental results based on feature extractors, architectures, standard datasets, and evaluation metrics are also presented to elaborate the performance of various models. 




\subsubsection{Our contributions:}
\label{contribution}
All these literature surveys provide a review on only a specific domain of multimodal learning. Some authors only discussed methods and applications of representation learning and some on fusion learning. Others discussed representation and fusion learning methods together using few modalities like vision or text. As mentioned earlier, Mogadala et al. \cite{mogadala2019trends} explained different models of MMDL using image, video and text modalities. However, we have taken one step further; along with these modalities, this paper provides a technical review of various models using audio, body gestures, facial expressions and physiological signals modalities. Our primary focus is distinctive in that we seek to survey the literature from up-to-date deep learning concepts using more modalities. The main contributions of our article are listed below:
\begin{table}[]
	\centering
	\scriptsize
	\caption{Analysis of baseline literature surveys, Where I=Image, T=Text, A=Audio, V=Video, BG=Body Gesture, FE=Facial Expression, PS=Physiological Signals, Med=DNA \& MRI}
	\label{tab:relatedSurveys}
	\begin{tabular}{|p{2cm}p{0.5cm}p{0.6cm}llllllllp{5cm}|}
		\bottomrule
		\multirow{2}{*}{\cellcolor{blue!25}} & \multirow{2}{*}{\cellcolor{blue!25}} & \multirow{2}{*}{\cellcolor{blue!25}} & \multicolumn{8}{c}{\textbf{\cellcolor{blue!25}Modalities}} & \multirow{2}{*}{\cellcolor{blue!25}} \\
		
		\textbf{\cellcolor{blue!25}Paper} & \textbf{\cellcolor{blue!25}Year} & \textbf{\cellcolor{blue!25}Pub.} & \textbf{\cellcolor{blue!25}I} & \textbf{\cellcolor{blue!25}T} & \textbf{\cellcolor{blue!25}A} & \textbf{\cellcolor{blue!25}V} & \textbf{\cellcolor{blue!25}BG} & \textbf{\cellcolor{blue!25}FE} & \textbf{\cellcolor{blue!25}PS} & \textbf{\cellcolor{blue!25}Med} & \textbf{\cellcolor{blue!25}Applications}  \\
		\bottomrule
		C. Zhang et al. \cite{zhang2020multimodal} & 2020 & IEEE & \checkmark & \checkmark &  & \checkmark &  &  &  & & Image Description, Image-to-Text generation, VQA  \\
		Y. Bisk et al. \cite{bisk2020experience} & 2020 & arXiv & \checkmark & \checkmark & \checkmark & \checkmark &  & & &  & Not Available  \\
		J. Gao et al. \cite{gao2020survey} & 2020 & NC(MIT) & \checkmark & \checkmark & \checkmark & \checkmark &  & &  &  & Not Available  \\
		A. Mogadala et al. \cite{mogadala2019trends} & 2019 & arXiv & \checkmark & \checkmark &  & \checkmark &  & & &  & Caption Generation, Storytelling, QA. Dialogue, Reasoning, Referring Expression, Entailment, Visual Generations, Navigation, Machine Translation  \\
		SF Zhang et al. \cite{zhang2019multimodal} & 2019 & IEEE & \checkmark & \checkmark & \checkmark & \checkmark &  &  & &  & Multimodal matching, Multimodal classification, Multimodal interaction, Multimedia content indexing, Multimedia Content retrieval, Multimedia description  \\
		W Guo et al. \cite{guo2019deep} & 2019 & IEEE & \checkmark & \checkmark & \checkmark & \checkmark &  &  & &  & Video Classification, Event Detection, VQA, Text-to-image synthesis, Transfer Loading  \\
		T Baltrušaitis et al. \cite{baltruvsaitis2018multimodal} & 2019 & IEEE & \checkmark & \checkmark & \checkmark & \checkmark &  & &  &  & Speech recognition, Media Description, Multimedia Retrieval  \\
		Y Li et al. \cite{li2018survey} & 2019 & IEEE & \checkmark & \checkmark & \checkmark & \checkmark &  &  & &  & Cross Media Retrieval, NLP, Video Analysis, Recommended System  \\
		D Ramachandram et al. \cite{ramachandram2017deep} & 2017 & IEEE & \checkmark & \checkmark & \checkmark & \checkmark &  &  & & \checkmark & Human Activity Recognition, Medical Applications, Autonomous Systems\\
		Ours & - & - & \checkmark & \checkmark & \checkmark & \checkmark & \checkmark & \checkmark & \checkmark & & Multimodal Image Description, Multimodal Video Description, Multimodal VQA, Multimodal Speech Synthesis, Multimodal Emotion Recognition, Multimodal Event Detection \\
		\bottomrule
	\end{tabular}
\end{table}

\begin{itemize}
	\item We propose a novel fine-grained taxonomy of various MMDL applications, which elaborates different groups of applications in more depth.
	\item We provide an explicit overview of various models using the modalities of images, videos, text, audio, body gestures, facial expressions, and physiological signals.
	\item We analyze and summarize the recent baseline MMDL models with the perspective of architecture, multimedia, dataset and evaluation metric used along with model features, which provide substantial guidance for future research inquiries.
	\item We provide a comparative analysis of MMDL models on benchmark datasets and evaluation metrics. Comparative analysis unique representation helps researchers to get directions in future research to improve domain performance using specific or combination of feature extractors and architectures.
	\item We highlights the main challenges of MMDL applications, open research problems, and their possible future research directions.
	
\end{itemize}

\subsection{Structure of Survey:}

\begin{figure}
	\centering
	\tmpframe{\includegraphics[width=\textwidth,height=60mm]{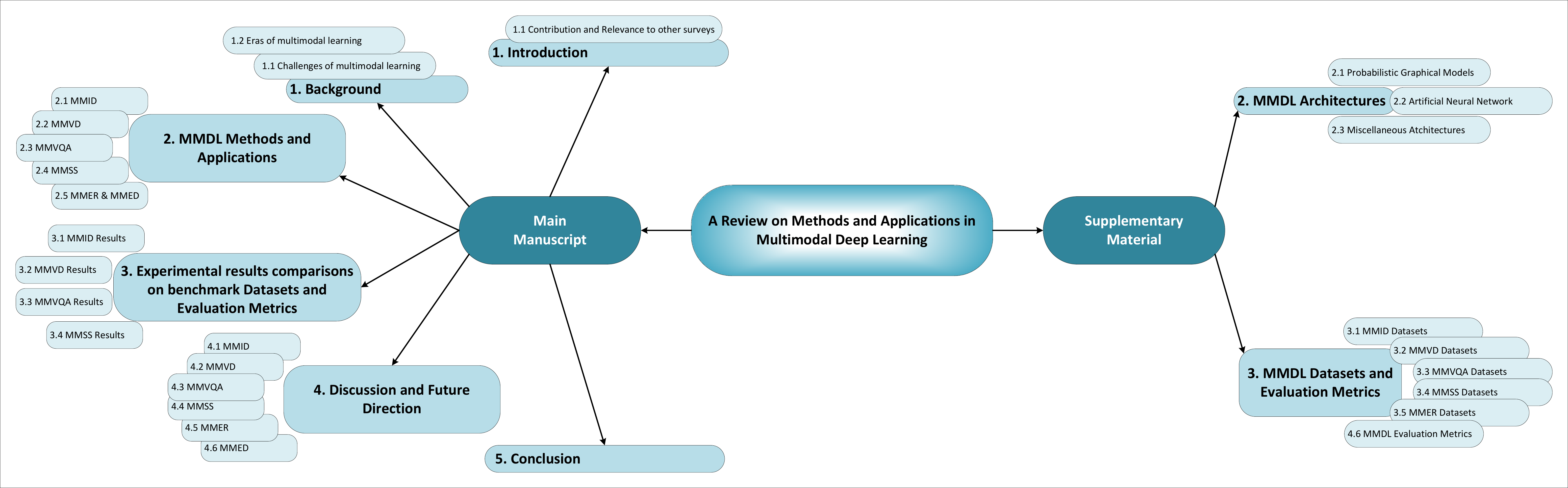}}
	\caption{Structure of Survey article.}
	\label{fig:structure of paper}
\end{figure}

The rest of the article is structured as follows. Some background of MMDL is discussed in section \ref{backGround}. Section \ref{AppsMMDL} presents various MMDL methods and applications. Comparative analysis on experimental results of MMDL application is discussed in section \ref{reslults}. In section \ref{Disc&future}, we provided a brief discussion and propose future research directions in this active area. Finally, section \ref{conclusion} concludes the survey paper. Due to page limits details of related MMDL architecture, and Dataset and evaluation metrics are provided in Section 1\footnote{Section 1 in supplementary meterials contains the complete details about architectures used in MMDL methods.}, and 2\footnote{Section 2 in supplementary materials contains the details of datasets \& evaluation metrics used in MMDL methods.} of supplementary material. Structure of survey article is shown in Figure \ref{fig:structure of paper}. 

\section{Background:} 
\label{backGround}

A lot of ML approaches have been developed over the past decade to deal with multimodal data. Multimodal machine learning leads to a wide range of applications: from Audio-visual speech recognition, multimedia content indexing and retrieval, understanding human multimodal behaviors, emotion recognition, multimodal affect recognition, image and video captioning, VQA, multimedia retrieval, to health analysis and so on. A list of abbreviations is presented in the Appendix A. 

Multimodal applications came into existence in the 1970s and were categorized into four eras \cite{Morency}. This section discusses the core challenges of multimodal learning, and the description of four eras is explained further. In this paper, we concentrate mainly on the deep learning era for multimodal applications.

\subsection{Challenges of multimodal learning:}
\label{challenges}

The research domain of MMDL presents some specific challenges for researchers due to heterogeneous data. Learning from multiple sources allows recording correspondences among various modalities and an in-depth comprehension of natural phenomena. In multimodal learning, extracting and combining features from multiple sources contributes to making larger-scale predictions. Data from various sources are contextually linked and occasionally give supplementary information to one another, revealing patterns that would not be noticeable when functioning with individual modalities separately. Such systems combine disparate, heterogeneous data from multiple sensors, allowing for more accurate predictions. Five main technical challenges of multimodal deep learning are summarized in this section, which is listed below.

\begin{itemize}
	\item Multimodal Representation: The process of representing information from several media in a tensor or vector form is known as multimodal representation. As information from several media frequently contains both redundant and complementary data, the goal is to represent data in a meaningful and efficient manner. Dealing with missing data, various levels of noise, and data combinations from several media are all challenges linked with multimodal representation. Joint and Coordinated representations are used to deal with these representation challenges.
	\item Multimodal Translation: The problem of translating or mapping information from one modality to another is addressed by multimodal translation. Consider the tasks like producing an image from a caption or producing a caption for an image. Due to data heterogeneity, it is not possible to create one perfect description of an image. One of the most challenging aspects of multimodal translation is assessing translation quality. The results quality for speech synthesis and video or image description is highly subjective, and there is frequently no single correct translation.
	\item Multimodal Alignment: The process of developing correspondence among information from two or more media for the same event is known as multimodal alignment. A framework must measure similarities among different media and cope with long-term dependencies to align them. Other challenges in the alignment task include the formation of better similarity metrics, the lack of annotated datasets, and various correct alignments.
	\item Multomodal Fusion: The process of combining data from two or more media to perform regression or classification is known as multimodal fusion. Sentiment analysis is one of the commonly known examples of multimodal fusion;  three modalities (Visual, acoustic, and language) combine to predict sentiment.
	\item Multimodal Co-learning: The process of transmitting information/knowledge among modalities is addressed by multimodal co-learning. Transmitting information/knowledge from a resource-rich modality is very effective for creating a model in a limited resource modality with noisy inputs, a lack of annotated data, and unreliable labels. Parallel, non-parallel, and hybrid co-learning approaches are used to transfer knowledge between modalities. 
\end{itemize}

\begin{figure}[t]
	\centering
	\tmpframe{\includegraphics[width= \textwidth, height=50mm]{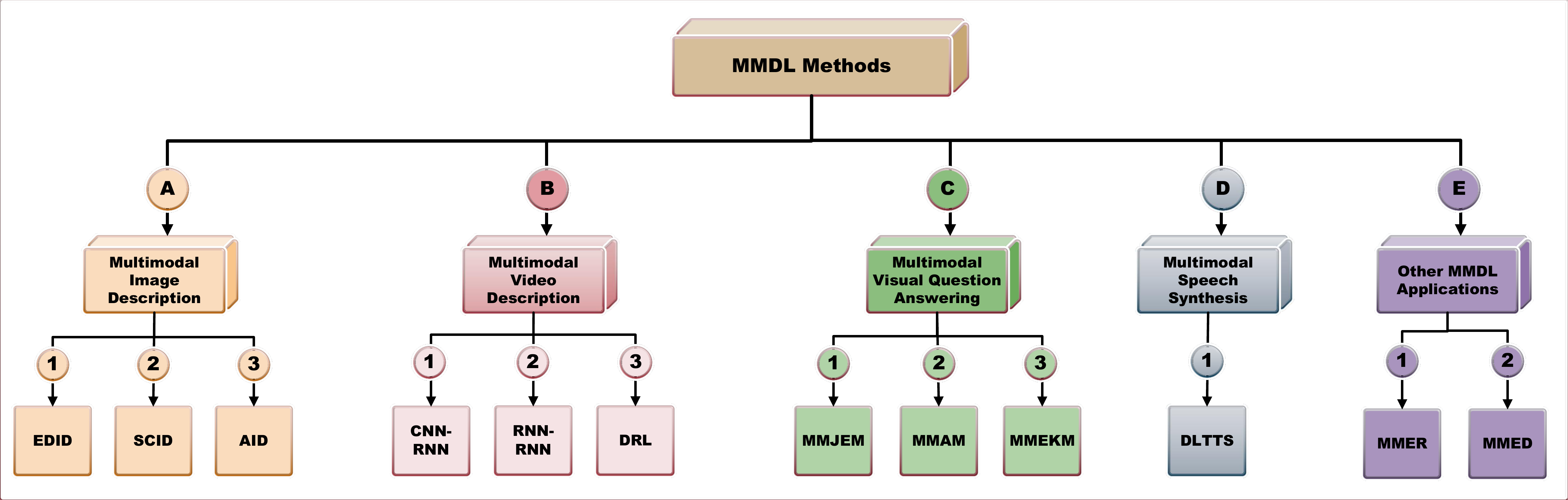}}
	\caption{Taxonomy diagram of Multimodal Deep Learning Applications. Where MMDL=Multimodal Deep Learning, EDID=Encoder-Decoder based Image Description, SCID=Semantic Concept-based Image Description, AID=Attention-based Image Description, DRL=Deep Reinforcement Learning, MMJEM=Multimodal Joint-Embedding Models, MMAM=Multimodal Attention-based Models, MMEKM=Multimodal External Knowledge Bases Models, DLTTS=Deep Learning Text To Speech, MMER=Multimodal Event Recognition, and MMED=Multimodal Emotion Detection.}
	\label{fig:appsTexonomy}
\end{figure}

\subsection{Eras of multimodal learning:}
\label{4eras}

Prior research on multimodal learning can be categorized into four eras, listed below.

\begin{itemize}
	\item Behavioral Era (BE): 
	The BE starts from the 1970s to the late 1980s. During this era, A Lazarus \cite{lazarus1973multimodal} proposed Multimodal behavior therapy based on seven different personality dimensions using inter-related modalities. 
	M. Mulligan \& M. Shaw \cite{mulligan1980multimodal} presents multimodal signal detection, wherein the signal detection task, the pooling of information is examined from other modalities, i.e., auditory and visual stimuli.
	L.E. Bahrick \cite{bahrick1983infants} found that infants detected bimodal temporal structure specifying object elasticity and rigidity and temporal synchrony between sights and sound of object. 
	D. Hoffman-Plotkin \cite{hoffman1984multimodal} proposed a ``Multimodal assessment of behavioral and cognitive deficits in abused and neglected pre-schoolers''. 
	
	\item Computational Era (CE): 
	The CE starts from the late 1980s to 2000. During this era, 
	E.D. Petajan \cite{petajan1985automatic} proposed an ``Automatic Lipreading to Enhance Speech Recognition.''
	B.P. Yuhas et al. \cite{yuhas1989integration} increase the performance of automatic speech recognition system even in noisy environments by using neural networks (NN).
	The inspiration for these approaches was McGurk effect \cite{mcgurk1976hearing} -- an interaction between vision and hearing during understanding of speech. 
	Hidden Markov Model (HMM) \cite{juang1991hidden} has become so popular in speech recognition because of the inherent statistical framework, the ease and accessibility of training algorithms to estimate model parameters from finite speech data sets, flexibility of the resulting recognition system, in which the size, type or architecture of the models can be easily changed to suit specific words, sounds.
	
	\item Interaction Era (IE):
	The IE starts from 2000 to 2010. During this era the understanding of human multimodal behaviors is achieved during social interactions. One of the first milestones is the AMI Meeting Corpus \cite{carletta2005ami} (a multimodal dataset) with over 100 hours of meeting recordings, all thoroughly annotated and transcribed. AMI seeks to create a repository of meetings and to evaluate conversations. 
	CHIL project motive \cite{waibel2009computers} uses ML methods to extract nonverbal human behaviors automatically. Under CHIL, different people tracking systems were built using audio, video or both modalities.
	CALO Meeting Assistant (CALO-MA) \cite{tur2008calo} real-time meeting recognition system takes the speech models. CALO-MA architecture includes real-time and offline speech transcription, action item recognition, question-answer pair identification, decision extraction, and summarization.
	Social Signal Processing (SSP) \cite{pantic2011social} aims at understanding and modelling social interactions and offering similar capabilities to computers in human-computer interaction scenarios.
	
	\item Deep Learning Era (DLE): 
	The DLE starts in 2010 to date. This deep learning era is the main focus of our review paper. We comprehensively discuss different groups of methods and applications proposed during this era in section \ref{AppsMMDL}, and datasets and evaluation metrics in section 2 of supplementary material proposed during this era. 
\end{itemize}

\section{MMDL Methods and Applications:}
\label{AppsMMDL}

Various methods and applications are designed using multimodal deep learning techniques. In this article, these methods and applications are grouped with relevance and dominance across multiple research areas. The taxonomy diagram of these applications is presented in the Figure \ref{fig:appsTexonomy}.

\subsection{Multimodal Image Description (MMID):}
\begin{figure}
	\centering
	\tmpframe{\includegraphics[width=160mm,height=70mm]{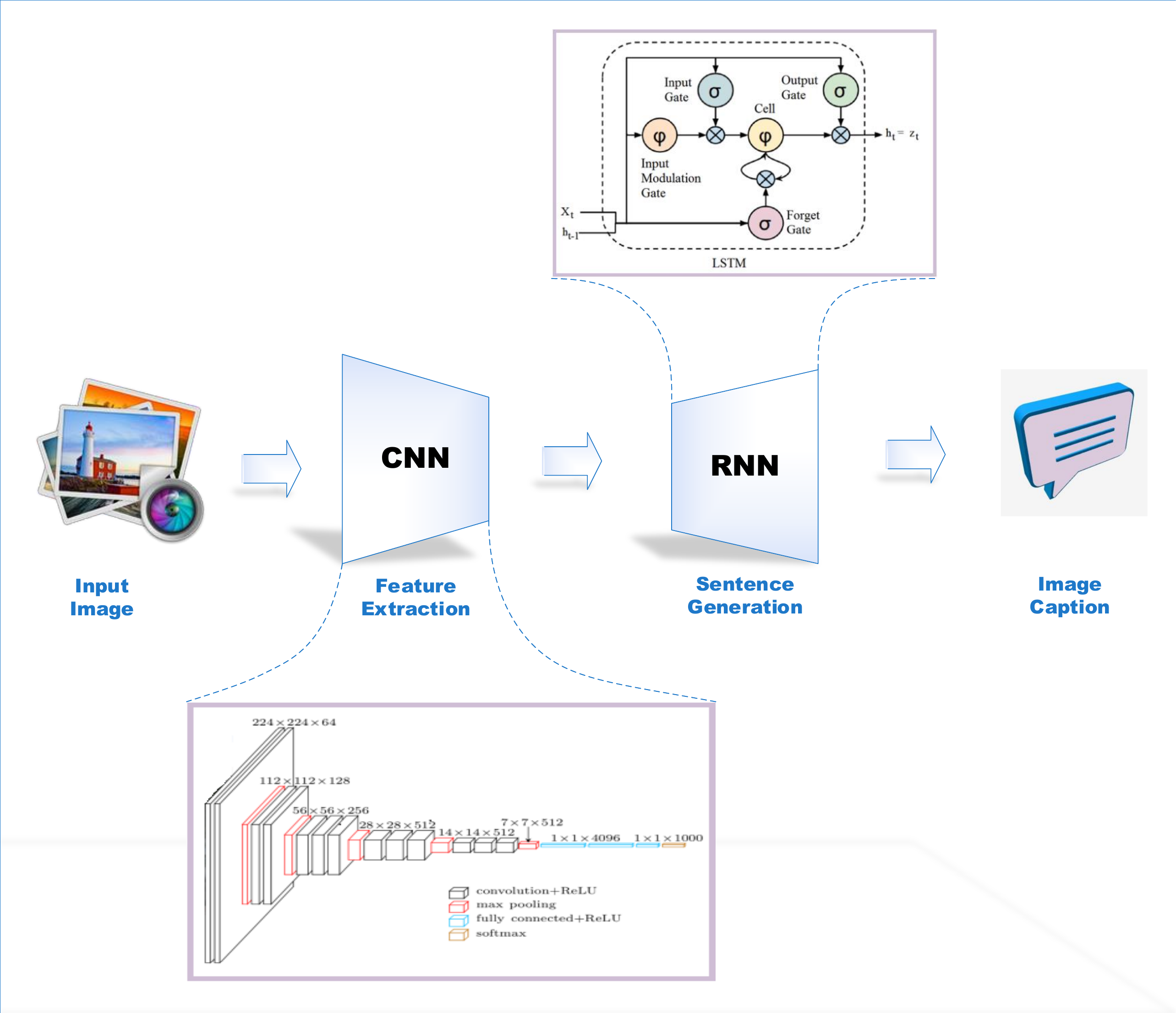}}
	\caption{General structure diagram of Image Description Model.}
	\label{fig:imgDescription}
\end{figure}

Image Description is mostly used to generate a textual description of visual contents provided through input image. During the deep learning era, two different fields are merged to perform image descriptions, i.e., CV and NLP. In this process, two main kinds of modalities are used, i.e., image and text. The image description's general structure diagram is shown in Figure \ref{fig:imgDescription}. Image description frameworks are categorized into Retrieval-based, Template-based, and DL-based image descriptions. Retrieval and Template based image descriptions are one of the earliest techniques for describing visual contents from images. In this article, DL-based image description techniques are explained in detail, which are further categorized into encoder-decoder-based, semantic concept-based, and attention-based image descriptions. Different image description approaches are analyzed comparatively according to architectures, multimedia, publication year, datasets, and evaluation metrics in Table \ref{tab:ImageDes}.

\subsubsection{Encoder-Decoder based Image Description (EDID):}
\label{EDID}
EDID plays a vital role in image captioning tasks using DL architectures. CNN architectures are used mainly as encoder parts to extract and encode data from images, and RNN architectures are used as decoder part to decode and generate captions.
J. Wu \& H.hu \cite{wu2017cascade} proposed a Cascade Recurrent Neural Network (CRNN) for image description. For the learning of visual language interactions, a cascade network is adopted by CRNN from the forward and backward directions. 
M. Chen et al. \cite{chen2017reference} proposed referenced based LSTM model for image description task. In this model, training images are used as reference for proposed framework, to minimize the misrecognitions for the description task.    
W. Jiang et al. \cite{jiang2018recurrent} proposed a recurrent fusion network for the task of image captioning on the basis of encoder-decoder. In this network, CNN architecture are used to extract information from input image and RNN architecture is used to generate the description in the form of text. 
L. Guo et al. \cite{guo2019mscap} proposed a multi-style image captioning framework using CNN, GAN, LSTM and GRU architectures. In this framework, five different captioning styles are introduced for the image: romantic, negative, positive, factual and humorous styles.
X. He et al. \cite{he2019image} proposed an image caption generation framework with the guidance of Part of Speech (PoS). In the word generation part, PoS tags are fed into LSTM as guidance to generate more effective image captions. 
Y. Feng et al. \cite{feng2019unsupervised} proposed an unsupervised image captioning framework. In this model first attempt is made to do captioning of the image without any labeled image-sentence pairs.
J. Ji et al. \cite{ji2021improving} introduced a globally enhanced transformation network for encoder and decoder. In this network, the encoder is used to extract global features from the image and the global adaptive controller at the decoder side is used for controlled image description.

\subsubsection{Semantic Concept-based Image Description (SCID):}
\label{SCID}
A collection of semantic concepts extracted from the image are selectively addressed by SCID approaches. These concepts are extracted at the encoding stage along with other features of an image, then merged into hidden states of language models, and output is used to generate descriptions of images based on semantic concepts. 
W. Wang et al. \cite{wang2018novel} proposed an attribute-based image caption generation framework. Visual features are extracted by using salient semantic attributes and are pass as input to LSTM’s encoder.
Z. Zhang et al. \cite{zhang2018high} proposed a semantic guided visual attention mechanism based image captioning model. Fully Convolutional Network (FCN) is primarily intended for semantic segmentation, especially for dense pixel level feature extractions and semantic labeling in the form of spatial grid. 
P. Cao et al. \cite{cao2019image} proposed a semantic-based model for image description. In this model, semantic attention-based guidance is used for LSTM architecture to produce a description of an image.  
L. Cheng et al. \cite{cheng2020stack} proposed a multi-stage visual semantic attention mechanism based image description model. In this approach, top-down and bottom-up attention modules are combined to control the visual and semantic level information for producing fine-grained descriptions of the image.
L. Chen et al. \cite{chen2021human} proposed a model to improve the accuracy of image captioning by introducing Verb-specific Semantic Roles (VSR). This model targets the activity and entities roles involved in that particular action to extract and generate the most specific information from image.

\begin{table}[]
	\scriptsize 
	\centering
	\caption{Comparative analysis of Image Description models. Where I=Image, T=Text, EM = Evaluation Metrics, B=BLEU, C=CIDEr, M=METEOR, R=ROUGE, S=SPICE}
	\label{tab:ImageDes}
	\begin{adjustbox}{width=1.1\textwidth, center=\textwidth}
		\begin{tabular}{||p{0.4cm}|p{1.3cm}|p{0.2cm}|p{3cm}|p{0.4cm}|p{1.7cm}|p{0.8cm}|p{10cm}|}
			\bottomrule
			\multicolumn{1}{|l|}{}&\multicolumn{1}{l|}{\textbf{Model}}&\multicolumn{1}{l|}{\textbf{Year}}&\multicolumn{1}{l|}{\textbf{Architecture}}& \multicolumn{1}{l|}{\textbf{Media}} & \multicolumn{1}{l|}{\textbf{Dataset}}&\multicolumn{1}{l|}{\textbf{EM}}&\multicolumn{1}{l|}{\textbf{Model Features}}    \\ 
			\bottomrule

			\multirow{7}{*}{} \cellcolor{orange!25} & CRNN \cite{wu2017cascade} & 2017 & CNN/VGG16-InceptionV3, SGRU & I,T & MS-COCO & B,C,M & \textbf{*} Cascaded network for learning visual language interactions. \textbf{*} Two embedding layers for dense word expressions are designed. \textbf{*} SGRU is designed for image to word mapping \\
			\cellcolor{orange!25}  & R-LSTM \cite{chen2017reference} & 2017 & CNN/VGG16, RNN/LSTM & I,T & MS-COCO & B,C, M,$R_{L}$ & \textbf{*} Trained images are used as reference for model \textbf{*} Words are weighted according to relevance score during training \textbf{*} Consensus score ranks the reference information for caption generation \\ 
			\cellcolor{orange!25}  & RFNet \cite{jiang2018recurrent} & 2018 & CNN/ResNet/DenseNet/ InceptionV3-V4,RNN/LSTM & I,T & MS-COCO & B,C,M, S,$R_{L}$ & \textbf{*} Use 5 feature extraction techniques \textbf{*} Multiple encoders are used for feature extraction \textbf{*} Comprehensive information/thought vectors are produced by fusing encoder information\\
			\cellcolor{orange!25} EDID & MSCap \cite{guo2019mscap} & 2019 & Deep CNN, GAN, RNN/LSTM, GRU & I,T & FlickrStyle10K, SentiCap,COCO & B,C,M, PPLX & \textbf{*} A unified multi-style caption generation model is introduced \textbf{*} Real human-like and stylized (Factual, Humorous, Romantic, Positive, Negative)captions are generated\\ 
			\cellcolor{orange!25}  & He et al \cite{he2019image} & 2019 & CNN/VGG16, 
			RNN/LSTM & I,T & Flickr30k, MS-COCO & B,C,M & \textbf{*} Use PoS sentence tags as guidance to LSTM word generator \textbf{*} PoS tags are controlled by switches to embed only specific information\\
			\cellcolor{orange!25}& Feng et al. \cite{feng2019unsupervised} & 2019 & CNN/InceptionV4, RNN/LSTM & I,T & MS-COCO & B,C,M, S,R & \textbf{*} "Visual Concept detector" provides guidance for caption generation \textbf{*} Visual feature and generated  captions are projected to a common latent space \textbf{*} Training is done without any labeled (image, sentence) pairs\\
			\cellcolor{orange!25}& GET \cite{ji2021improving} & 2021 & Faster-RCNN/ResNet, RNN/LSTM & I,T & MS-COCO & B,C,M, S,$R_{L}$& \textbf{*} Global gated adaptive controller is used to fuse relavent information at decoder \textbf{*} Inter and Intra layer representations are used to merge local/global information\\
			\bottomrule
			
			\multirow{4}{*}{} \cellcolor{red!15} & Wang et al. \cite{wang2018novel} & 2018 & CNN/VGG16,
			RNN/LSTM &I,T & MS-COCO & B,C,M & \textbf{*} Feature extraction using semantic attributes enhance the caption generation accuracy \textbf{*} Multiple instance learning transfer rules are designed at fc7 layer of VGG-16 to extract more semantic features\\
			\cellcolor{red!15} & FCN-LSTM \cite{zhang2018high} & 2018 & FCN/VGG16, RNN/LSTM & I,T & MSCOCO, COCO-Stuff & B,C,M & \textbf{*} Semantic-guided attention mechanism enhance the caption generation accuracy \textbf{*} Model works well for large areas like sky, beach, etc. by assigning additional semantic context \\
			\cellcolor{red!15} SCID & Bag-LSTM \cite{cao2019image} & 2019 & CNN/VGG16, RNN/BLSTM & I,T & Flickr8k, MS-COCO& B,C,M & \textbf{*} Feedback propagation extracts the text-related image features \textbf{*} Semantic attention-based guidance leverage the information dynamically \\
			\cellcolor{red!15} & Stack-VS \cite{cheng2020stack} & 2020 & Faster-RCNN, RNN/LSTM & I,T& MS-COCO & B,S, M,C,R & \textbf{*} Multi-Stage Bottom-Up and Top-Down stacked attention generates rich captions \textbf{*} Two LSTM layers at decoder re-optimize the semantic attention weights\\
			\cellcolor{red!15} & VSR \cite{chen2021human} & 2021 & Faster-RCNN, SSP, RNN/LSTM & I,T & Flickr30K, MS-COCO & B,C,M, S,$R_{L}$ & \textbf{*}Verb-specific Semantic Roles controls the image caption generation process \textbf{*}Human-like semantic structures ranks the verb \& semantic roles using SSP\\
			
			\bottomrule
			
			\multirow{8}{*}{} \cellcolor{blue!15} & GLA \cite{li2017gla} & 2017 & CNN/VGG16-Faster RCNN, RNN/LSTM & I,T & Flickr8K/30K, MS-COCO & B,C,M, $R_{L}$ &  \textbf{*} Object-level features are extracted along with image level features \textbf{*} Important features are extracted by focusing on specific regions of image using global-local attentions\\ 
			\cellcolor{blue!15}& Up-Down \cite{anderson2018bottom}  & 2018 & Faster RCNN/ResNet101, RNN/LSTM, GRU &I,T &  MS-COCO & B,C,M, S,R & \textbf{*} Top-down attention mechanism ranks extracted features from image \textbf{*} LSTM one layer is designed as top-down attention model and other layer as language model\\
			\cellcolor{blue!15}& NICVATP2L \cite{liu2020chinese} & 2020 & CNN/InceptionV4, RNN/LSTM & I,T & Flickr8k-CN, AIC-ICC & B,C,M, R & \textbf{*} Topic modeling introduced to increase the accuracy of caption generation \textbf{*} Visual regions and topic features are merged at decoding stage to guide two-layers LSTM for image captioning\\
			\cellcolor{blue!15} AID & NICNDA \cite{liu2020image} & 2020 & CNN/InceptionV4, RNN/LSTM & I,T & AIC-ICC & B,C,M, R & \textbf{*} Visual attention at decoder increases the image understanding \textbf{*} Information integrity is boosted by textual attention \textbf{*} Language model and textual attention are work together to fine-grain description\\
			\cellcolor{blue!15}& Wang et al. \cite{wang2020cross} & 2020 & CNN/InceptionV4, RNN/LSTM & I,T & Flickr8K, Flickr8k-CN & B, R, C, M & \textbf{*} Attention mechanism based on vector similarity is introduced \textbf{*} Semantic words and image features correlation express the appropriate attention \\
			\cellcolor{blue!15}& MAGAN \cite{wei2020multi} & 2020 & GAN, RNN/LSTM &I,T & MS-COCO & B,C,M, S,R & \textbf{*} Local and non-local evidences are used to rich the feature representations \textbf{*} Multi-attention discriminator and generator increased the image caption accuracy\\
			\cellcolor{blue!15}& MGAN \cite{jiang2021multi} & 2021 & Faster-RCNN, RNN/LSTM & I,T & MS-COCO & B,C,M, $R_{L}$,S & \textbf{*} Self Gate and Attention Weight Gate are used with existing self-attention mechanism to extract intra-object relationships \textbf{*} Pre-layernorm transformer is designed for features enhancement.\\
			
			\bottomrule
		\end{tabular}
	\end{adjustbox}
\end{table}

\subsubsection{Attention-based Image Description (AID):}
\label{AID}
AID plays a vital role because it helps the image description process by focusing on distinct regions of the image according to their context. In recent years, various techniques have been proposed to better describe an image by applying an attention mechanism. Some of these attention mechanism based image descriptions techniques are;   
L. Li et al. \cite{li2017gla} proposed a new framework for describing images by using local and global attention mechanisms. Selective object-level features are combined with image-level features according to the context using local and global attention mechanisms. 
P. Anderson et al. \cite{anderson2018bottom} proposed bottom-up and top-down attention based framework for image description to encourage deeper image understanding and reasoning. 
M. Liu et al. proposed a dual attention mechanism-based framework to describe an image for Chinese  \cite{liu2020chinese} and English \cite{liu2020image} languages. The textual attention mechanism is used to improve the data credibility, and the visual attention mechanism is used to a deep understanding of image features.
B. Wang et al. \cite{wang2020cross} proposed an E2E-DL approach for image description using a semantic attention mechanism. In this approach, features are extracted from specific image regions using an attention mechanism for producing corresponding descriptions.
Y. Wei et al. \cite{wei2020multi} proposed an image description framework by using multi attention mechanism to extract local and non-local feature representations.    
W. Jiang et al. \cite{jiang2021multi} proposed a multi-gate expansion of self-attention mechanism. In this network, attention mechanism is expanded by adding self-gated module and attention weight gate module to eliminate the irrelevant information from description.  

\begin{figure}
	\centering
	\tmpframe{\includegraphics[width=\textwidth,height=60mm]{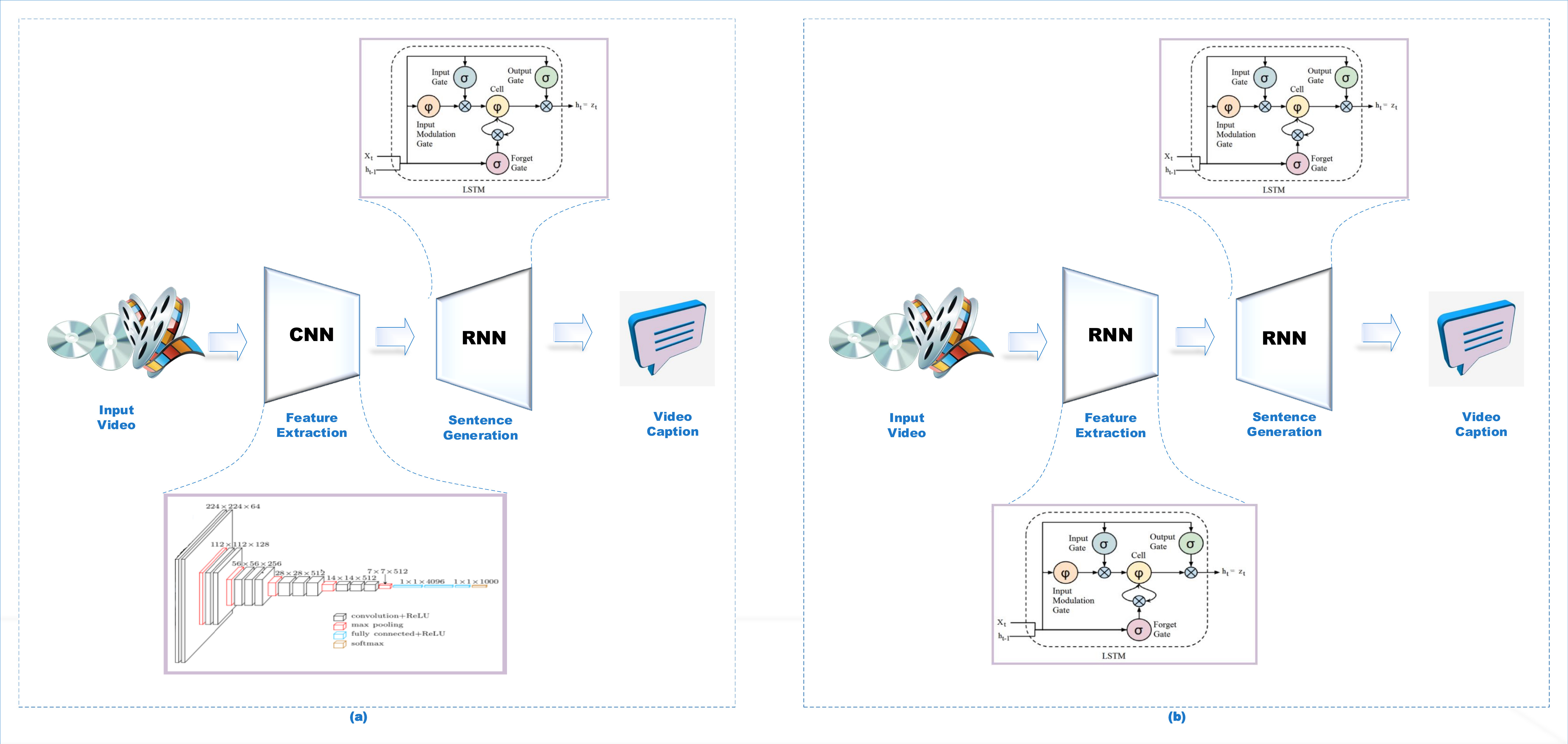}}
	\caption{General structure diagram of Video Description Model. (a) is general structure diagram of CNN-RNN architecture video description and (b) is General structure diagram of RNN-RNN architecture combination for video description.}
	\label{fig:VidDescription}
\end{figure}

\subsection{Multimodal Video Description (MMVD):}
\label{MMVDisc}
Like image Description, video description is used to generate a textual description of visual contents provided through input video. Here, DL approaches for the description of visual contents from videos are discussed in detail. Advancements in this field open up many opportunities in various application domains. During this process, mainly two types of modalities are used, i.e., video stream and text. The general structure diagram of the video description is shown in Figures \ref{fig:VidDescription} (a) \& (b). 
In this research, the video description approach is categorized based on the following architectural combinations for visual feature extraction and text generation. These approaches are comparatively analyzed according to architectures, multimedia, datasets, and evaluation metrics in Table \ref{tab:VidDescSummary}.

\subsubsection{CNN-RNN Architectures:}
\label{CNN-RNN VidD}
Most broadly used architecture combination in the domain of video description is CNN-RNN. Figure \ref{fig:VidDescription}(a) presents the general view of the video description process by using CNN-RNN architectures. Where at the visual extraction (encoder) stage variants of CNN architectures are used and at the sentence generation (decoder) stage variants of RNN architectures are used. During deep learning era, several authors proposed techniques for describing videos that are based on this encoder, decoder combination. 
R Krishna et al. \cite{krishna2017dense} proposed a video description technique using action/event detection by applying a dense captioning mechanism. This is the first framework to detect and describe several events, but it didn't significantly improve video captioning.
B. Wang et al. \cite{wang2018reconstruction} proposed a reconstruction network for video description using an encoder-decoder reconstructor architecture, which utilizes both forward flow (from video to sentence) and backward flow (from sentence to video).  
W. Pei et al. \cite{pei2019memory} proposed an attention mechanism based encoder-decoder framework for video description. An additional memory based decoder is used to enhance the quality of video description.
N Aafaq et al. \cite{aafaq2019spatio} proposed video captioning and capitalized on Spatio-temporal dynamics of videos to extract high-level semantics using 2D and 3D CNNs hierarchically, and GRU is used for the text generation part.
S. Liu et al. \cite{liu2020sibnet} proposed SibNet; a sibling convolutional network for video description. Two architectures are used simultaneously to encode video, i.e., the content-branch to encode visual features and the semantic-branch to encode semantic features. 
J. Perez-Martin et al. \cite{perez2021improving} improves the visual captioning quality by implementing visual syntactic embedding. A PoS tagging structure is used to extract the syntactic representations from video and guide the decoder with these temporal compositions to achieve accurate descriptions.

\begin{table}[] 
	\scriptsize
	\centering
	\caption{Comparative analysis of Video description models. where, DRL=Deep Reinforcement Learning, V=Video, T=Text, EM = Evaluation Metrics, B=BLEU, C=CIDEr, M=METEOR, R=ROUGE }
	\label{tab:VidDescSummary}
	\begin{adjustbox}{width=1.1\textwidth, center=\textwidth}
		\begin{tabular}{||p{0.4cm}|p{1.5cm}|p{0.2cm}|p{3cm}|p{0.4cm}|p{2cm}|p{0.8cm}|p{10cm}|}
			\bottomrule
			\multicolumn{1}{|l|}{}&\multicolumn{1}{l|}{\textbf{Model}}&\multicolumn{1}{l|}{\textbf{Year}}&\multicolumn{1}{l|}{\textbf{Architecture}}& \multicolumn{1}{l|}{\textbf{Media}} & \multicolumn{1}{l|}{\textbf{Dataset}}&\multicolumn{1}{l|}{\textbf{EM}}&\multicolumn{1}{l|}{\textbf{Model Features}}    \\ 
			\bottomrule
			
			\multirow{9}{*}{\cellcolor{orange!30}}  & Krishna et al. \cite{krishna2017dense} & 2017 & CNN/C3D, RNN/LSTM & V,T & ActivityNet Captions & B,C,M & \textbf{*} Dense event detection and description are introduced \textbf{*} All the events are identified in one pass then forward to language model for rich description \\
			\cellcolor{orange!30}   & RecNet \cite{wang2018reconstruction} & 2018 & CNN/InceptionV4, RNN/LSTM & V,T & MSVD,   MSR-VTT, & B,C,M   $R_{L}$ & \textbf{*} Both video to sentence and sentence to video flows are monitered \textbf{*} Video to sentence flow generates the description with video semantic guidance \\
			\cellcolor{orange!30} CNN to & MARN \cite{pei2019memory} & 2019 & CNN/ResNet101/ResNeXt101, RNN/GRU & V,T & MSVD, MSR-VTT & B,C,M, $R_{L}$ & \textbf{*} A memory structure explores the correlation between words and its visual features during training \textbf{*} Description quality is improved by this correlation \textbf{*} Modeling of adjacent words are done explicitly\\
			\cellcolor{orange!30} RNN & GRU-MP/EVE \cite{aafaq2019spatio} & 2019 & CNN/IRV2, 3DCNN/C3D, RNN/GRU & V,T & MSVD, MSR-VTT & B,C,M, R & \textbf{*} Temporal instances in visual features are extracted by hierarchically applying Short Fourier Transform \textbf{*} Semantic features are extracted during object detection to enhance visual representation \\
			\cellcolor{orange!30}& SibNet \cite{liu2020sibnet} & 2020 & CNN/GoogLeNet/Inception, RNN/LSTM & V,T & MSVD, MSR-VTT & B,C,M, R & \textbf{*} Two branch (context and semantic) modules simultaneously encode visual features \textbf{*} Both branches features are merged with soft-attention for rich and concise descriptions\\
			\cellcolor{orange!30}& SemSynAN \cite{perez2021improving} & 2021 & 2D-CNN/3D-CNN, RNN/LSTM & V,T & MSVD, MSR-VTT & B,C,M, $R_{L}$ & \textbf{*} Syntactic representation extracts the rich visual information \textbf{*} SemSynAN guide language model by merging syntactic, visual and semantic representations to enhance description accuracy\\
			
			\bottomrule
			\multirow{4}{*}{\cellcolor{green!15}} & SSVC \cite{rahman2020semantically} & 2020 & CNN/VGG16-LSTM, RNN/LSTM & V,T & MSVD & B & \textbf{*} Content generation technique is modified using stacked attention and spatial hard pull \textbf{*} Objects in videos are prioritized layer-by-layer using stacked attention.\\
			\cellcolor{green!15} RNN to& V2C \cite{fang2020video2commonsense} & 2020 & CNN/ResNet-LSTM, RNN/LSTM & V,T & V2C,   MSR-VTT & B,M, $R_{L}$ & \textbf{*} Commonsense captions are generated for events detected from videos \textbf{*} V2C-transfer mechanism generates the commonsense enriched captions\\
			\cellcolor{green!15} RNN & GPaS \cite{zhang2020dense} & 2021 & GCN, RNN/LSTM, RNN/LSTM & V,T & ActivityNet Captions & B,C,M $R_{L}$ & \textbf{*} Partition and Summarization strategy enhances the video description \textbf{*} Partition splits event proposal into video segments and summarization module generates description for each segment\\
			
			\bottomrule
			\multirow{7}{*}{\cellcolor{blue!18} } 
			{\cellcolor{blue!18} }	& HRL \cite{wang2018video} & 2018 & CNN/ResNet, RNN/BLSTM/LSTM, HRL &V,T & MSR-VTT, Charades & B,C,M, $R_{L}$ & \textbf{*} Manager and Worker modules design sub-goals and compulsory actions to accomplish these goals \textbf{*} Attention mechanism focus widely on temporal dynamics of visual features\\
			{\cellcolor{blue!18} }	& PickNet \cite{chen2018less} & 2018 & RNN/LSTM, RNN/GRU & V,T & MSVD,   MSR-VTT & B,C,M, $R_{L}$ & \textbf{*} Pick selective informative frames from video to extract visual features \textbf{*} Video description can be done with compact frames \textbf{*} Computation cost of image description is reduced\\
			{\cellcolor{blue!18} }	& E2E. \cite{li2019end} & 2019 & CNN/Inception-Resnet-V2, RNN/LSTM & V,T & MSVD,   MSR-VTT & B,C,M, $R_{L}$ & \textbf{*} During training, attributes extracted from videos are used to maximize the reward \textbf{*} High-level features information is extracted using five duplicates networks of CNN at encoder\\
			{\cellcolor{blue!18} DRL}	& SDVC \cite{mun2019streamlined} & 2019 & SST, C3D, RNN/GRU/LSTM & V,T & ActivityNet Captions & B,C,M & \textbf{*} From videos, temporal dependencies are extracted explicitly \textbf{*} Linguistic and visual features context are extracted from previous events are core part of video description process\\
			{\cellcolor{blue!18} }	& RecNet(RL) \cite{zhang2019reconstruct} & 2019 & CNN/InceptionV4, RNN/LSTM & V,T & ActivityNet, MSVD, MSR-VTT & B,C,M, $R_{L}$ & \textbf{*} A similar approach to RecNet for video description task with addition of RL technique to improve  description \textbf{*} Two kinds of reconstructors extract local and global representations from video\\
			{\cellcolor{blue!18} }	& DPRN \cite{xu2020deep} & 2020 & MDP, RNN/LSTM & V,T & MSVD, MSR-VTT & B,C,M, $R_{L}$ & \textbf{*} Word denoising and grammar checking networks refine the generated descriptions \textbf{*} Long term reward enhances the description by considering global representations into account\\
			{\cellcolor{blue!18} }	& Wei et al. \cite{wei2020exploiting} & 2020 & 3D CNN/ResNet, RNN/LSTM & V,T & MSVD,   MSR-VTT, Charades & B,C,M & \textbf{*} Adaptive moving window extracts better event representations \textbf{*} Additionally, temporal attention enhance description process by focusing on temporal frame features while words captioning\\
			\bottomrule
			
		\end{tabular}
	\end{adjustbox}
\end{table}

\subsubsection{RNN-RNN Architectures:}
\label{Rnn-Rnn VidD}
During the DL era, RNN-RNN is also a popular architectural combination because many authors contribute a lot by proposing various methods using this combination. Authors extract the visual content of the video by using RNN architectures instead of CNN. Figure \ref{fig:VidDescription}(b) presents the general view of the video description process by using RNN-RNN architectures. Both visual extraction (encoder) and sentence generation (decoder) stage variants of RNN architectures are used.  
M. Rahman Et al. \cite{rahman2020semantically} proposed a video captioning framework that modifies the generated context using spatial hard pull and stacked attention mechanisms. This approach illustrates that mounting an attention layer for a multi-layer encoder will result in a more semantically correct description. 
Z. Fang et al. \cite{fang2020video2commonsense} proposed a framework to generate commonsense captions of the input video. Commonsense description seeks to identify and describe the latent aspects of video like effects, attributes, and intentions. A new dataset, "Video-to-Commonsense (V2C)," is also proposed for this framework.
Z. Zhang et al. \cite{zhang2020dense} proposed an encoder decoder based framework for dense video captioning. A graph based summarization and partition modules are used to enhance the word relation ship between context and event fount in video.

\subsubsection{Deep Reinforcement Learning (DRL) architectures:}
\label{DRL VidD}
\begin{figure}
	\centering
	\tmpframe{\includegraphics[width=\textwidth,height=60mm]{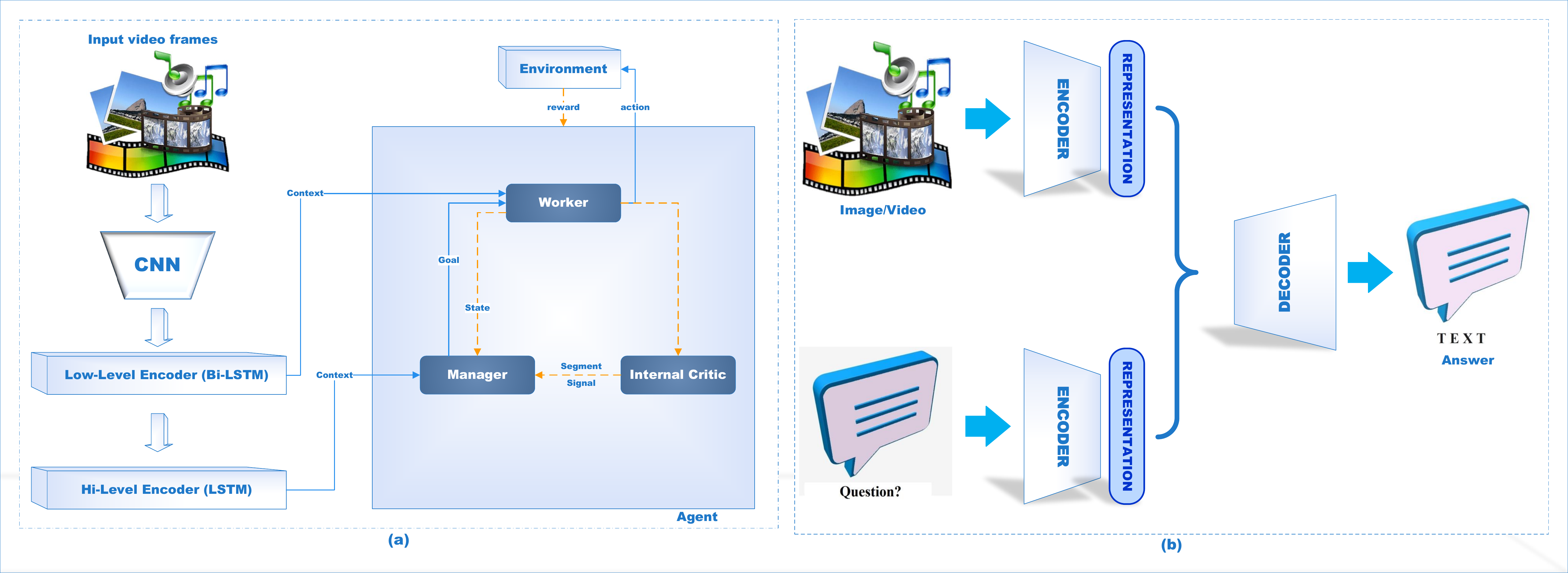}}
	\caption{(a) General structure diagram of Video Description Deep Reinforcement Learning Architectures, and (b) General structure diagram of VQA System}
	\label{fig:VidDescRL and VQA}
\end{figure}

DRL is a learning mechanism where machines can learn intelligently from actions like human beings can learn from their experiences. In it, an agent is penalized or rewarded based on actions that bring the model closer to the target outcome. The general structure diagram of the video description DRL is shown in Figure \ref{fig:VidDescRL and VQA} (a). The main contributions of authors using DRL architectures are;
X. Wang et al. \cite{wang2018video} proposed a hierarchical based reinforcement learning (HRL) model for describing a video. In this framework, a high-level manager module designed sub-goals and low-level worker module recognize actions to fulfill these goals.  
Y. Chen et al. \cite{chen2018less} proposed a framework based on RL for choosing informative frames from an input video. Fewer frames are required to generate video description in this approach.
L. Li \& B. Gong \cite{li2019end} proposed an E2E multitask RL framework for video description. The proposed method combines RL with attribute prediction during the training process, which results in improved video description generation.
J. Mun et al. \cite{mun2019streamlined} proposed a framework where an event sequence generation network is used to monitor the series of events for generated captions from the video.  
W. Zhang et al. \cite{zhang2019reconstruct} proposed a reconstruction network for a description of visual contents, which operates on both forward flow (from video to sentence) and backward flow (from sentence to video).
W. Xu et al. \cite{xu2020deep} proposed a polishing network that utilizes the RL technique to refine the generated captions. This framework consists of word denoising and grammar checking networks for fine-tuning generated sentences. 
R. Wei et al. \cite{wei2020exploiting} proposed a framework for better exploration of RL events to generate more accurate and detailed video captions. 

\subsection{Multimodal Visual Question Answering (MMVQA):} 
\label{MMVQA}
VQA is an emerging technique that has piqued the interest of both the CV and NLP groups. It is a field of research about creating an AI System capable of answering natural language questions. Extracted features from input image/video and question are processed and combined to answer the question about the image as presented in Figure \ref{fig:VidDescRL and VQA} (b).
VQA is more complex as compared to other vision and language functions like text-to-image retrieval, video captioning, image captioning, etc. because; (1) Questions asked in VQA are not specific or predetermined. (2) Visual information in VQA is at a high degree of dimensionality. Usually, VQA required a more thorough and detailed understanding of an image/video. (3) VQA solves several CV sub-tasks. Many authors contribute to the field of VQA by using various DL techniques. These methods are grouped and presented into three groups, i.e., multimodal joint-embedding models, multimodal attention-based models, and multimodal external knowledge-based models. Various methods of these models are comparatively analyzed in Table \ref{tab:VQASummary}.

\subsubsection{Multimodal Joint-Embedding Models (MMJEM):}
\label{VQAJoint}

MMJEM join and learn representations of multiple modalities in a common feature space. This rationale is improved further in VQA by performing more reasoning over modalities than image/video description.  
H. Ben-Younes et al. \cite{ben2017mutan} proposed a MUTAN framework for VQA. A tensor-based tucker decomposition model is used with a low-rank matrix constraint to parameterize the bi-linear relations between visual and text interpretations.  
MT Desta et al. \cite{desta2018object} proposed a framework that merges the visual features and language with abstract reasoning. High-level abstract facts extracted from an image optimize the reasoning process. 
R. Cadene et al. \cite{cadene2019murel} proposed an E2E reasoning network for VQA. This research's main contribution is introducing the MuRel cell, which produces an interaction between question and corresponding image regions. 
B. Patro et al. \cite{patro2020robust} proposed a joint answer and textual explanation generation model. A collaborative correlated (encoder, generator, and correlated) module is used to ensure that answer and its generated explanation is correct and coherent.  
S. Lobry et al. \cite{lobry2020rsvqa} proposed a VQA framework for remote sensing data, which can be useful for land cover classification tasks. 
Z. Fang et al. \cite{fang2020video2commonsense} proposed an open-ended VQA framework for videos using commonsense reasoning in the language part, where questions are asked about effects, intents, and attributes. 

\subsubsection{Multimodal Attention-based Models (MMAM):}
\label{VQAAttention}
During the encoding stage, a general encoder-decoder can feed some noisy and unnecessary information at the prediction phase. MMAM are designed to improve the general baseline models to overcome this problem. The attention mechanism's main objective is to use local features of image/video and allow the system to assign priorities to the extracted features from different regions. This concept is also used in VQA to enhance model's performance by focusing on a specific part of an image according to the asked question.

P. Wang et al. \cite{wang2017vqa} proposed a VQA framework based on a co-attention mechanism. In this framework, the co-attention mechanism can process facts, images, and questions with higher order. 
Z. Yu et al. \cite{yu2017multi} proposed a factorized bi-linear pooling method with co-attention learning for the VQA task. Bilinear pooling methods outperform the conventional linear approaches, but the practical applicability is limited due to their high computational complexities and high dimensional representations.
P. Anderson et al. \cite{anderson2018bottom} proposed bottom-up and top-down attention based framework for VQA. This attention mechanism enables the model to calculate feature based on objects or salient image regions. 
Z. Yu et al. \cite{yu2019deep} proposed a deep modular co-attention network for the VQA task. Each modular co-attention layer consists of question guided self-attention mechanism of images using the modular composition of both image and question attention units. 
L. Li et al. \cite{li2019relation} proposed a relation aware graph attention mechanism for VQA. This framework encodes the extracted visual features of an image into a graph, and to learn question-adaptive relationship representations, a graph attention mechanism modeled inter-object relations.  
W. Guo et al. \cite{guo2020re} proposed a re-attention based mechanism for VQA. The attention module correlates the pair of object-words and produces the attention maps for question and image with each other's guidance. 

\begin{table}[]
	\scriptsize
	\centering
	\caption{Comparative analysis of Visual Question Answering models. where MMJEM=Multimodal Joint-Embedding Models, MMAM=Multimodal Attention-based Models, MMEKM=Multimodal External Knowledge bases Models, I=Image, V=Video, T=Text, EM=Evaluation Metrics,  Acc=Accuracy, B=BLEU, C=CIDEr, M=METEOR, R=ROUGE, S=SPICE}
	\label{tab:VQASummary}
	\begin{adjustbox}{width=1.1\textwidth, center=\textwidth}
		\begin{tabular}{||p{0.7cm}|p{1.5cm}|p{0.2cm}|p{3cm}|p{0.4cm}|p{1.7cm}|p{0.8cm}|p{10cm}|}
			\bottomrule
			\multicolumn{1}{|l|}{}&\multicolumn{1}{l|}{\textbf{Model}}&\multicolumn{1}{l|}{\textbf{Year}}&\multicolumn{1}{l|}{\textbf{Architecture}}& \multicolumn{1}{l|}{\textbf{Media}} & \multicolumn{1}{l|}{\textbf{Dataset}}&\multicolumn{1}{l|}{\textbf{EM}}&\multicolumn{1}{l|}{\textbf{Model Features}}    \\ 
			\bottomrule
			
			\multirow{11}{*}{\cellcolor{orange!30}} & MUTAN \cite{ben2017mutan} & 2017 & CNN/ResNet152, RNN/GRU & I,T & VQA & Acc & \textbf{*} Bilinear interactions between textual and visual features are effectively parameterize by tucker framework \textbf{*} Matrix-based module is designed to explicitly constrain the interaction rank\\
			\cellcolor{orange!30} & Desta et al. \cite{desta2018object} & 2018 & Faster RCNN/Resnet101, RNN/LSTM, RNN/GRU & I,T & CLEVR & Acc & \textbf{*} Reasoning efficiency is increased by extracted abstract facts from text and video inputs \textbf{*} Visual features extraction relies on information linked with objects \\
			\cellcolor{orange!30}  & MuReL \cite{cadene2019murel} & 2019 & Faster RCNN, GRU & I,T & VQAv2.0,TDIUC, VQA-CPv2 & Acc & \textbf{*} MuReL cell represents correlation between image and question regions \textbf{*} Cell features are incorporated into whole model to refine question and visual representations\\
			\cellcolor{orange!30} \tiny \textbf{MMJEM} & CCM \cite{patro2020robust} & 2020 & CNN, RNN/LSTM & I,T & VQA-X & B,S,C, M,$R_{L}$ & \textbf{*} A collaborative correlated module verifies the coherency of explanations and answers \textbf{*} Attention maps used in these correlated networks are closely related with human generated maps \\
			\cellcolor{orange!30}& RSVQA \cite{lobry2020rsvqa} & 2020 & CNN/ResNet152, RNN & I,T & OpenStreetMap (LR,HR) & Acc & \textbf{*} RSVQA system extracts features from remote sensing data \textbf{*} High-level features are associated with relational dependencies among visible objects\\
			\cellcolor{orange!30}& V2C \cite{fang2020video2commonsense} & 2020 & CNN/ResNet-LSTM, RNN/LSTM & V,T & V2C-QA & TopK & \textbf{*} Open ended commonsense reasoning is used to enhance the VQA quality \textbf{*} This reasoning part also helps to improve the video captioning task \\
			
			\bottomrule
			\multirow{9}{*}{\cellcolor{green!15}} &  Wang et al. \cite{wang2017vqa}& 2017 & CNN/VGG19-ResNet100, RNN/LSTM & I,T & Visual Genome QA, VQA-real & WUPS, Acc & \textbf{*} Different combination of CV algorithms learns adaptively generate the reasoning part \textbf{*} Existing Co-Attention model is upgraded to process images, facts and question jointly \\
			\cellcolor{green!15}  & MFB \cite{yu2017multi} & 2017 &	CNN/ResNet152, RNN/LSTM & I,T & VQA MS-COCO & Acc & \textbf{*} MFB framework fuse the textual and visual features from question and images jointly \textbf{*} And co-attention structure learns the question and image attentions\\
			\cellcolor{green!15} &  Up-Down \cite{anderson2018bottom} & 2018 & Faster R-CNN, RNN/GRU & I,T & Visual Genome, VQA v2.0 & Acc & \textbf{*} Top-down attention mechanism ranks extracted features from image and used question features as context \textbf{*} Multimodal embedding of image and question increase the VQA model accuracy\\
			\cellcolor{green!15} \tiny \textbf{MMAM} & MCAN \cite{yu2019deep} & 2019 & Faster R-CNN/ResNet101, RNN/LSTM & I,T & VQA v2.0 & Acc & \textbf{*} "Modular Co-Attention (MCA)" layer use composition of images and questions self-attention \textbf{*} MCA layer gradually refines the question and image representations\\
			\cellcolor{green!15}  & ReGAT \cite{li2019relation} & 2019 & Faster R-CNN/ResNet101, RNN/LSTM  & I,T & VQA v2.0, VQA-CP v2 & Acc & \textbf{*} Both implicit and explicit relations are used to get appropriate image and question representations \textbf{*} ReGAT captures/records the inter-object relations to reveal fine-grained features from image\\
			\cellcolor{green!15} & Xi et al. \cite{xi2020visual} & 2020 & CNN, RNN/LSTM  & I,T & DQAUAR, COCO-QA & Acc, WUPS & \textbf{*} Appearance model is extended by word vector similarity to enhance VQA accuracy \textbf{*} Appearance features form objects are used instead of image features \\
			\cellcolor{green!15} & Guo et al. \cite{guo2020re} & 2020 & Faster R-CNN, RNN/LSTM  & I,T & VQA v2.0 & Acc & \textbf{*} Merge image and question features are associated together to find object similarities \textbf{*} Then according to generated answer, model re-attend the associated object and reconstruct attention maps to improve answers \\

			\bottomrule
			\multirow{7}{*}{\cellcolor{yellow!18} } &  FVQA \cite{wang2018fvqa} & 2018 & Faster R-CNN/VGG16, RNN/LSTM  & I,T & MS-COCO, FVQA & TopK,Acc, WUPS & \textbf{*} The dataset designed for VQA task based on facts \textbf{*} These facts are gathered from external sources (DBPedia, ConceptNet, Web Child) to produce answers \\
			{\cellcolor{yellow!18}}	& Narasimhan et al. \cite{narasimhan2018straight} & 2018 & CNN/Faster RCNN/VGG16, RNN/LSTM & I,T & FVQA & Acc, TopK & \textbf{*} Parametric mapping of image-questions pairs and facts are gathered into a common embedding space \textbf{*} Most aligned facts are used to produce answer of question\\
			{\cellcolor{yellow!18} } \tiny \textbf{MMEKM} & OK-VQA \cite{marino2019ok} & 2019 & CNN/ResNet, RNN/GRU & I,T & OK-VQA & Recall, TopK & \textbf{*} OK-VQA is a diverse, large and also a difficult database as compared to existing databases \textbf{*} Ok-VQA works well where model requires external information instead of only visual features\\
			{\cellcolor{yellow!18}}	& GRUC \cite{yu2020cross} & 2020 & Faster-RCNN, RNN/LSTM & I,T & FVQA,OKVQA, Visual7W+KB  & Acc, TopK & \textbf{*} Image features are extracted by multiple knowledge graph from three (semantic, factual and visual) views \textbf{*} Recurrent reasoning model extract the compulsory representation from multi model space\\
			{\cellcolor{yellow!18} } & AQuA \cite{basu2020aqua} & 2020 & YOLO, SRE & I,T & CLEVR & Acc & \textbf{*} Answer set programming is introduced to effectively understand the query about image \textbf{*} To deal with complex queries, semantic features are extracted and used to enhance scenario understanding \\
			\bottomrule
			
		\end{tabular}
	\end{adjustbox}
\end{table}

\subsubsection{Multimodal External Knowledge Bases Models (MMEKM):}
\label{VQAknowledge}
Traditional multimodal joint embedding and attention-based models only learn from the information that is present in training sets. Existing datasets do not cover all events/activities of the real world. Therefore, MMEKM is vital to coping with real-world scenarios. Performance of VQA task is more increasing by linking knowledge bases (KB) databases to VQA task.  
Freebase \cite{bollacker2008freebase}, DBPedia \cite{auer2007dbpedia}, WordNet \cite{miller1995wordnet}, ConceptNet \cite{liu2004conceptnet}, and WebChild \cite{tandon2014webchild} are extensively used KB. A robust VQA framework requires access to broad information content from KB. It has been effectively integrated into the VQA task by embedding the various entities and relations. 

During the DL era, various external KB methods are proposed for VQA tasks. 
P. Wang et al. \cite{wang2018fvqa} proposed another framework for the VQA task names "Fact-based VQA (FVQA)" that uses data-driven approaches and LSTM architecture to map image/question queries. FVQA framework used DBPedia, ConceptNet, and WebChild KB. 
M. Narasimhan \& AG. Schwing \cite{narasimhan2018straight} proposed a framework for the VQA task using external knowledge resources that contain a set of facts. This framework can answer both fact-based and visual-based questions. 
K. Marino et al. \cite{marino2019ok} proposed an outside knowledge dataset for VQA, which contains more than 14,000 questions. This dataset contains several categories like sports, science and technology, history, etc. This dataset requires external resources to answer, instead of only understanding the question and image features.  
K. Basu et al. \cite{basu2020aqua} proposed a commonsense based VQA framework. In this framework, the image's visual contents are extracted and understood by the YOLO framework and represented in the answer set program.  Semantic relations features and additional commonsense knowledge answer the complex questions for natural language reasoning.
J. Yu et al. \cite{yu2020cross} proposed a framework in which visual contents of an image is extracted and processed in multiple perspectives of knowledge graph like semantic, visual, and factual perspectives. 

\begin{figure}
	\centering
	\tmpframe{\includegraphics[width=\textwidth,height=70	mm]{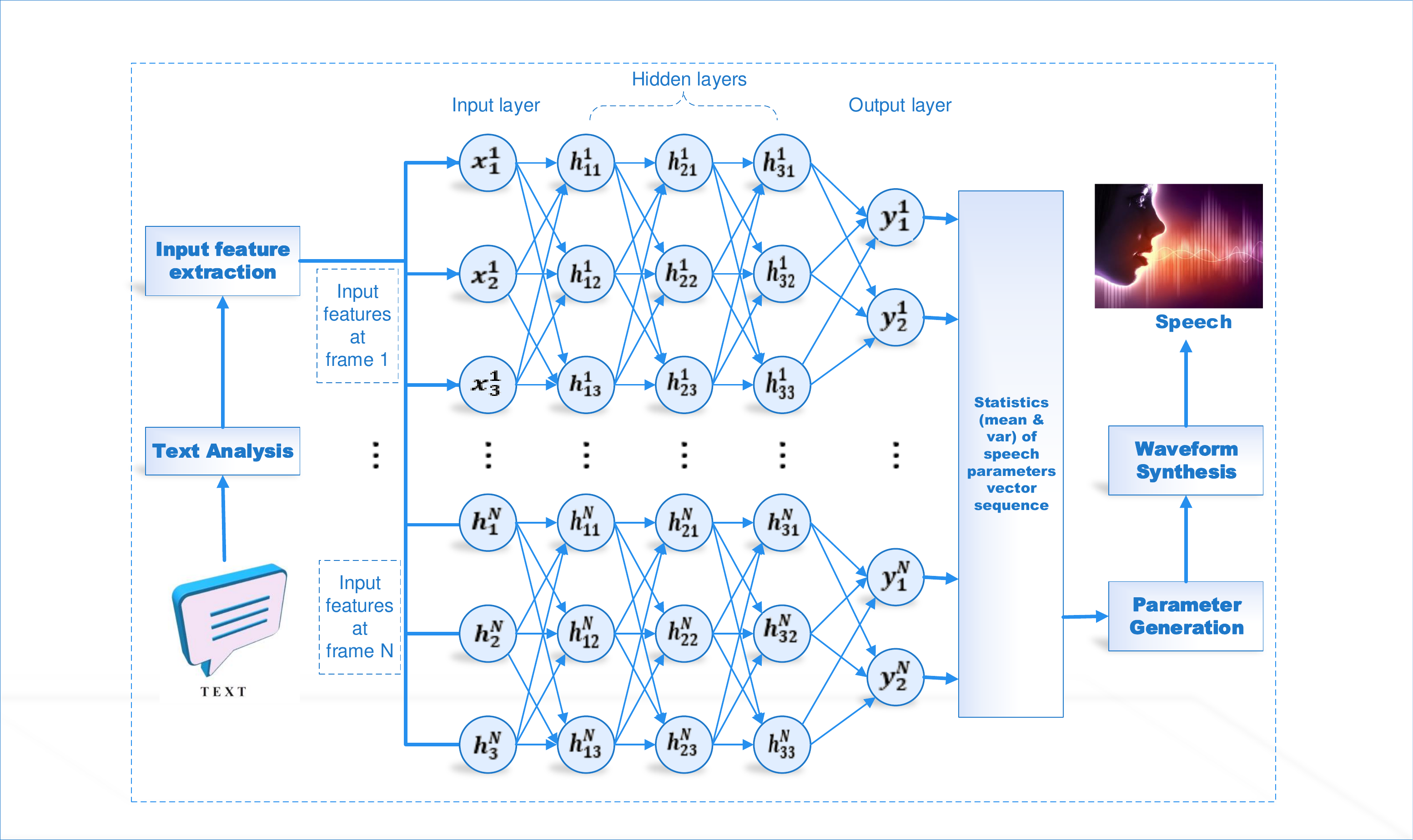}}
	\caption{General structure diagram of Deep Learning Text To Speech framework using DNN architecture.}
	\label{fig:SpeechSynthesis}
\end{figure}

\subsection{Multimodal Speech synthesis (MMSS):}
The most important aspect of human behavior is communication (write/speak). Humans can communicate using natural language by text and speech, representing the written and vocalized form of natural language, respectively. The latest research in language and speech processing helps systems talk like a human being. Speech synthesis is the complicated process of generating natural language spoken by a machine. Natural language text modality is converted into its respective spoken waveform modality in real-time by the Text To Speech (TTS) system. Various applications are introduced in the real world using speech synthesis, like human-computer interactive systems, screen readers, telecommunication and multimedia applications, talking toys games, etc. The main research objective of TTS systems these days is to produce a sound like a human's. Therefore, various aspects are used for the evaluation of the TTS system's quality, such as naturalness (quality with the perspective of generated speech timing structure, rendering emotions and pronunciation), intelligibility (in a sentence, quality of each word being produced), synthetic speech preferences (choice of a listener in term of voice and signal quality for better TTS system) and human perception factors like comprehensibility (understanding quality of received messages). Articulatory TTS, Concatenative TTS, Formant TTS, Parametric TTS, and Deep Learning TTS are the main categories of the speech synthesis process. This section discusses recent research trends and advancements of Deep Learning TTS.

\begin{table}
	\scriptsize
	\centering
	\caption{Comparatively analysis of speech synthesis (DLTTS) models, where T=Text, A=Audio, V=Video, EM=Evaluation Metrics, Acc=Accuracy }
	\label{tab:TTS}
	\begin{adjustbox}{width=1.1\textwidth, center=\textwidth}
		\begin{tabular}{||p{2.1cm}|p{0.6cm}|p{2cm}|p{0.5cm}|p{3.1cm}|p{0.9cm}|p{10cm}|}
			\bottomrule
			\cellcolor{blue!20}\textbf{Model}  & \cellcolor{blue!20}\textbf{Year} &  \cellcolor{blue!20}\textbf{Architecture}  & \cellcolor{blue!20}\textbf{Media} & \cellcolor{blue!20}\textbf{Dataset} & \cellcolor{blue!20}\textbf{EM} &\cellcolor{blue!20}\textbf{Model Features}\\
			\bottomrule
			TACOTRON \cite{wang2017tacotron} & 2017 & Bidirectional-GRU/RNN & T,A & North American English & MOS & \textbf{*} TACOTRON synthesizes speech directly from corresponding characters embedding \textbf{*} Frame-based technique increases the speed of inference procedure\\
			DeepVoice \cite{arik2017deep} & 2017 & GRU/RNN & T,A & Internal English speech database, Blizzard & MOS & \textbf{*} Five modules (Segmentation, grapheme-phoneme, phoneme duration prediction, frequency prediction, and audio synthesis) synthesize the speech from text \textbf{*} Inference speed of DeepVoice is faster than baseline methods \\
			DeepVoice2 \cite{gibiansky2017deep} & 2017 & GRU/RNN & T,A & Internal English speech database & MOS & \textbf{*} Improvement over DeepVoice by introducing higher performance building blocks and multi-speaker speech synthesis \textbf{*} Processing vocoder improves the quality of generated speech\\
			DeepVoice3 \cite{ping2017deep} & 2017 & Griffin-Lim, WaveNet vocoder & T,A & VCTK, Internal English speech database, LibriSpeech & MOS & \textbf{*} DeepVoice3 is upgraded with the addition of attention mechanism to improve speech quality \textbf{*} The training and scaling speed is also increases\\
			Parallel WaveNet \cite{oord2018parallel} & 2018 & CNN & A & North American English, Japanese & MOS & \textbf{*} To increase the speech synthesis processing speed, probability density distillation model is used \textbf{*} Parallel WaveNet is much faster than WaveNet and produce multiple voices\\
			Arik et al. \cite{arik2018neural} & 2018 & Griffin-Lim vocoder & T,A & LibriSpeech, VCTK & Acc,EER, MOS & \textbf{*} This model clone the person's voice with fewer audio samples \textbf{*} Speaker encoding and adaptation models infer new speaker embedding and fine tune these embeddings \\
			VoiceLoop \cite{taigman2018VoiceLoopVF} & 2018 & GMM & T,A & Lj speech, Blizzard, VCTK & MOS,MCD TopK & \textbf{*} Shifting buffer working memory estimates the attention and synthesize the audio \textbf{*} VoiceLoop deal un-constrained audio samples without phoneme alignment\\
			TACOTRON2 \cite{shen2018natural} & 2018 & STFT, BiLSTM & T,A & Internal US English & MOS & \textbf{*} Feature prediction model maps characters to mel-scale spectograms \textbf{*} A modified WaveNet model synthesize these spectograms to wave form\\
			AV-ASR \cite{tao2020end} & 2020 & CNN, RNN/LSTM & A,V & Manual data collection (CRSS-4ENGLISH-14 corpus) & CER, F-Score & \textbf{*} AV-ASR combines the modalities fusion feature extraction, classification task and information modeling to improve synthesis process \textbf{*} Multitask learning improves the framework performance\\
			Parallel TACOTRON \cite{elias2020parallel} & 2021 & GLU, VAE, LSTM & T,A & Proprietary speech & MOS & \textbf{*} Model is highly parallelizable during inference and training process \textbf{*} Auto-encoder improves naturalness by one-to-many mapping \\
			\bottomrule
		\end{tabular}
	\end{adjustbox}
\end{table}

\subsubsection{Deep Learning TTS (DLTTS):}
\label{DLTTS}
In DLTTS frameworks, DNN architectures model the relationship between text and their acoustic realizations. The main advantage of DLTTS is the development of its extensive features without human prepossessing. Also, the naturalness and intelligibility of speech are improved using these systems. Text to speech synthesis process is explained in the general structure diagram of Deep Learning Text To Speech frameworks using DNN architectures, shown in the Figure \ref{fig:SpeechSynthesis}. Comparative analysis of these approaches is shown in Table \ref{tab:TTS}. 

Y. Wang et al. \cite{wang2017tacotron} proposed "Tacotron," a sequence 2 sequence TTS framework that synthesizes speech from text and audio pairs. Encoder embeds the text that extracts its sequential representations. The attention-based decoder process these representations, and after that, post-processing architecture generates the synthesized waveforms.
In an another research, "Deep Voice" model using DNN architecture are proposed by SO Arik et al. \cite{arik2017deep} synthesizes audio from characters. This model consists of five significant blocks for the production of synthesized speech from text. The computational speed is increased compared to existing baseline models because the model can train without human involvement.
A. Gibiansky et al. \cite{gibiansky2017deep} proposed a Deep Voice-2 architecture. This framework is designed to improve existing state-of-the-art methods, i.e., Tacotron and Deep Voice-1, by extending multi-speaker TTS through low-dimension trainable speaker embedding.
In third version of Deep Voice, W. Ping et al. \cite{ping2017deep} proposed a neural TTS system based on the fully convolutional model with an attention mechanism. This model performs parallel computations by adapting Griffin-Lim spectrogram inversion, WORLD, and WaveNet vocoder speech synthesis.  
"Parallel WaveNet" an advanced version of WaveNet proposed by A. Oord et al. \cite{oord2018parallel} using the probability density distribution method to train networks. A teacher and a student WaveNet are used parallelly in this model. 
SO Arik et al. \cite{arik2018neural} proposed a neural voice cloning system that learns human voice from fewer samples. For that purpose, two techniques, i.e., speaker adaptation and encoding, are used together.
Y. Taigman et al. \cite{taigman2018VoiceLoopVF} proposed a VoiceLoop framework for a TTS system. This model can deal with un-constrained voice samples without the need for linguistic characteristics or aligned phonemes. This framework transformed the text into speech from voices using a short-shifting memory buffer.  
J. Shen et al. \cite{shen2018natural} proposed "Tacotron2. It is neural TTS architecture used to synthesize speech directly from the text. A recurrent based sequence to sequence feature prediction network can map characters to spectrogram, and then these spectrogram are used to synthesize waveforms by using a modified version of WaveNet vocoder.
F. Tao and C. Busso \cite{tao2020end} proposed an speech recognition system using multitask learning mechanism. The proposed design takes into account the temporal dynamics across and within modalities, resulting in an enticing and feasible fusion method.
Parallel Tacotron is another brilliant invention in recent times for neural TTS approach proposed by I. Elias et al. \cite{elias2020parallel}. During inference and training processes, this approach is highly parallelizable to achieve optimum synthesis on modern hardware. One to many mapping nature of VAE enhance the performance of TTS and also improves its naturalness.

\begin{figure}
	\centering
	\tmpframe{\includegraphics[width= \textwidth, height=70mm]{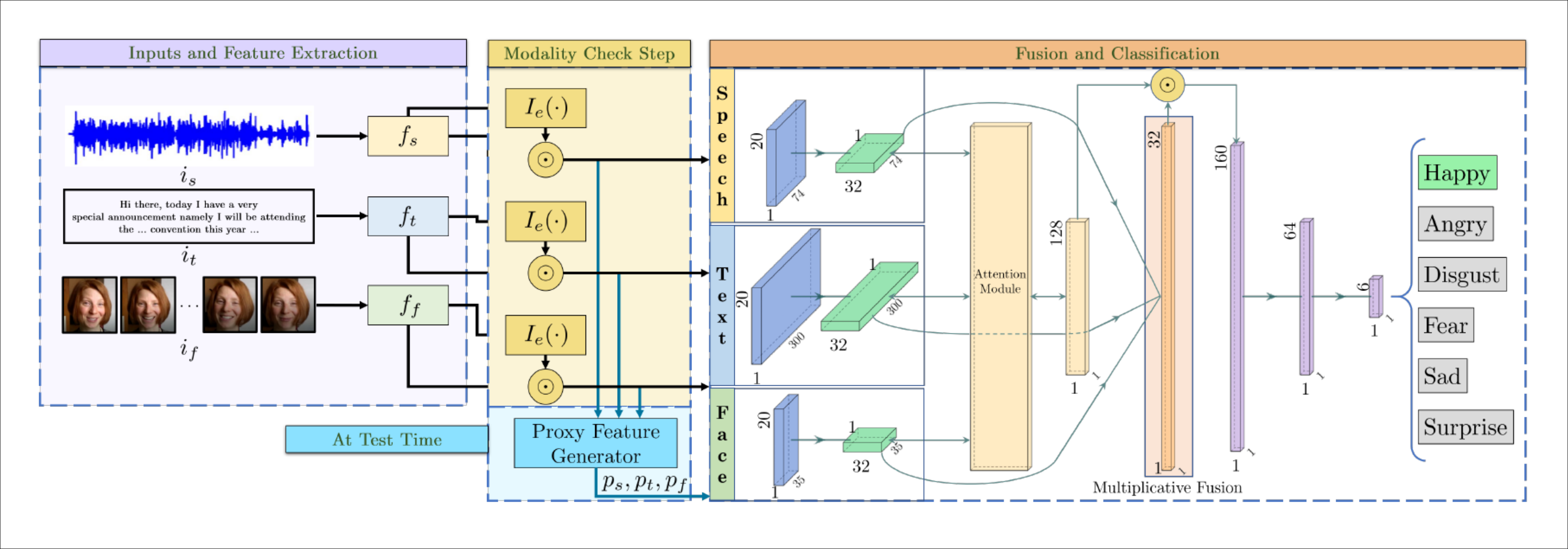}}
	\caption{General structure diagram of Multimodal Emotion Recognition using Facial, Textual, and Speech Cues \cite{mittal2020m3er}.}
	\label{fig:EmotionDetection}
\end{figure}

\subsection{Other MMDL Applications:}
\label{others apps}

\subsubsection{Multimodal Emotion Recognition (MMER):}
\label{EmotionRecog}

Emotions of human is one of the approach for expressing feelings. MMER is enormously vital for enhancing the interaction experience between humans and computers. The task of ML is to empower computers to learn and identify new inputs from training datasets, hence it can be used efficiently to detect, process, respond, understand and recognize emotions of human through computers training. Thus, the primary purpose of affective computing is to provide machines/systems emotional intelligence ability. It has several research areas such as learning, health, education, communication, gaming, personalized user interface, virtual reality and information retrieval. Multimodal emotion recognition framework can be developed on an AI/ML based prototypes designed to extract and process emotion information from various modalities like speech, text, image, video, facial expression, body gesture, body posture, and physiological signals. The general structure diagram of multiplicative multimodal emotion recognition using facial, textual, and speech cues are shown in Figure \ref{fig:EmotionDetection}.

During the DL era, various authors contribute to emotion recognition by using different architecture and multiple modalities. Like,    
Y. Huang et al. \cite{huang2017fusion} proposed a fusion method for emotion recognition using two modalities, i.e., facial expression and electroencephalogram (EEG) signals. A NN classifier detects happiness, neutral, sadness, and fear states of emotions.  
D. Nguyen et al. \cite{nguyen2017deep} proposed another approach for emotion recognition by using audio and video streams. In this method, a combination of C3D and DBN architectures is used to model Spatio-temporal information and representation of video audio streams.
D. Nguyen et al. \cite{nguyen2018deep} proposed another approach for emotion recognition by using audio and video streams. In this method, a combination of C3D, MCB and DBN architectures is used to model Spatio-temporal information and representation of video audio streams.
S. Tripathi et al. \cite{tripathi2018multi} proposed a multimodal emotion recognition framework using various modalities like, text, speech, face expression, and hands movement on the IEMOCAP dataset. The fusion of modalities is performed only at the final layer to improve emotion recognition performance.
D. Hazarika et al. \cite{hazarika2018icon} proposed an emotion recognition framework from video conversations. This method can generate self and interpersonal affective summaries from conversations by using contextual information extracted from videos. They also proposed another framework for detecting emotions using the attention mechanism \cite{hazarika2018self}. In this framework, audio and textual modalities are used for the detection of emotions.  
M. Jaiswal et al. \cite{jaiswal2019controlling} analyzes the change of emotional expressions under various stress levels of an individual. The performance of this task is affected by the degree of lexical or acoustic features.  
L. Chong et al. \cite{chong2019emochat} proposed a new online-chatting system called "EmoChat" to automatically recognize the user's emotions and attach identified emotion automatically and the sent message within a short period. During the chatting user can know each other's emotions in this network.
M. Li et al. \cite{li2020multistep} proposed a multi-step deep system for reliable detection of emotions by using collected data that contains invalid data as well. Neural networks are used to filter this invalid data from videos and physiological signals by using continuity and semantic compatibility.  
H. Lai et al. \cite{lai2020different} proposed a model for emotion recognition in interactive conversations. In this model, different RNN architectures with variable contextual window sizes differentiate various aspects of contexts in conversations to improve the accuracy. 
RH Huan et al. \cite{huan2020video} proposed a model using attention mechanism. In this method, for better use of visual, textual and audio features for video emotion detection, bidirectional GRU is cascaded with an attention mechanism. The attention function is used to work with various contextual states of multiple modalities in real-time.  
Y Cimtay et al. \cite{cimtay2020cross} proposed a hybrid fusion method for emotion recognition using three modalities, i.e., facial expression, galvanic skin response (GSR) and EEG signals. This model can identify the actual emotion state when it is prominent or concealed because of natural deceptive face behavior. 

In this section, recent existing methods for emotion recognition and analysis using multimodal DL are presented. Comparative analysis of these techniques is shown in Table \ref{tab:OthersSummary}. This literature review on emotion detection is based on text, audio, video, physiological signals, facial expressions and body gestures modalities. It is clearly shown from an empirical study that automatic emotion analysis is feasible and can be very beneficial in increasing the accuracy of the system response and enabling the subject's emotional state to be anticipated more rapidly. Cognitive assessment and their physical response are also analyzed along with primary emotional states. 

\begin{table}[]
	\scriptsize 
	\centering
	\caption{Comparative analysis of MMER and MMED models, Where PS=Physiological Signals, FE=Facial Expression, BE=Body Gesture I=Image, V=Video, A=Audio, T=Text, EM=Evaluation Metrics, WAcc=Weighted Accuracy, P=Precision, R=Recall}
	\label{tab:OthersSummary}
	\begin{adjustbox}{width=1.1\textwidth, center=\textwidth}
		\begin{tabular}{||p{0.6cm}|p{1.8cm}|p{0.2cm}|p{3cm}|p{0.4cm}|p{2cm}|p{1.2cm}|p{10cm}|}
			\bottomrule
			\multicolumn{1}{|l|}{}&\multicolumn{1}{l|}{\textbf{Model}}&\multicolumn{1}{l|}{\textbf{Year}}&\multicolumn{1}{l|}{\textbf{Architecture}}& \multicolumn{1}{l|}{\textbf{Media}} & \multicolumn{1}{l|}{\textbf{Dataset}}&\multicolumn{1}{l|}{\textbf{EM}}&\multicolumn{1}{l|}{\textbf{Model Features}}    \\ 
			\bottomrule
			
			\multirow{12}{*}{
				\cellcolor{red!20}} & Huang et al. \cite{huang2017fusion} & 2017 & SVM & PS,FE & Manual offline data collection & WAcc & \textbf{*} Fusion between peripheral and brain signals are used for emotion recognition \textbf{*} Three intensities (weak, moderate, and strong) are detected against each emotion\\
			\cellcolor{red!20}& A-V C3D+DBN \cite{nguyen2017deep} & 2017 & C3D, DBN & V,A & eNTERFACE Audio-visual & WAcc & \textbf{*} Three dimensional CNN model spatio-temporal representations from audio and video modality \textbf{*} Four combination of C3D and DBN models increases the accuracy of framework\\
			\cellcolor{red!20} &  A-V C3D+MCB +DBN \cite{nguyen2018deep} & 2018 & C3D, MCB, DBN & V,A &FABO,eNTERFACE Audio-visual & WAcc & \textbf{*} A-V C3D+DBN framework is further improved with the addition of bi-linear pooling \textbf{*} Feature-level fusion technique increases the accuracy of baseline method\\
			\cellcolor{red!20} & Tripathi et al. \cite{tripathi2018multi}&2018&MFCC, RNN/LSTM& A,T, FE,BE	&IEMOCAP&	WAcc & \textbf{*} Multiple modalities (text, speech and emotions captured from hand movement, face expressions) are used for emotion recognition task \textbf{*} Each modality is targeted individually and finally fuse together to get accurate representations\\
			\cellcolor{red!20}  & ICON \cite{hazarika2018icon}&2018&CNN, 3D-CNN, RNN/GRU& T,A,V &	IEMOCAP, SEMAINE& WAcc,MAE, F1-Score & \textbf{*} Multimodal representations are extracted from all utterance-videos \textbf{*} Contextual summaries for conversation are used for emotion recognition\\
			\cellcolor{red!20} & Hazarika et al. \cite{hazarika2018self}&2018&CNN, OpenSMILE, MFCC& T,A &	IEMOCAP &	WAcc,R, F1-Score & \textbf{*} Text and audio modalities are assigned proper score using attention mechanism \textbf{*} Prioritize features from these modalities are fuse together for accurate emotion recognition\\
			\cellcolor{red!20} MMER & Jaiswal et al. \cite{jaiswal2019controlling}& 2019 & GRU & PS, T & MuSE, IEMOCAP, MSP-Improv & WAcc & Adversial networks are designed to decorrelate stress representations from emotion features \textbf{*} Model shows stress representation affect the quality of emotion recognition \\
			\cellcolor{red!20}  & EmoChat \cite{chong2019emochat}&2019&TextCNN, Mini-Xception, HMM& T, FE 	&FER2013&	WAcc& \textbf{*} Emotions of user are identified online from text conversation \textbf{*} Joint processing of text and Facial expressions predicts the emotions in real-time\\
			\cellcolor{red!20}& MSD \cite{li2020multistep}&2020&C3D, DBN, SVM-RBF& V, PS &	RECOLA&	WAcc,R, F1-Score & \textbf{*} MSD process and filter the collected records containing invalid data \textbf{*} MSD emotion recognition performance improves automatically by extracting invalid data\\
			\cellcolor{red!20}& DCWS-RNNs \cite{lai2020different}&2020&CNN, 3D-CNN, RNN/GRU& T,A,V & IEMOCAP, AVEC	& Acc, F1-Score& \textbf{*} Four RNNs with different sizes context window is used to convert context into memories \textbf{*} Attention-based multiple maps merged these memories for emotion recognition\\
			\cellcolor{red!20}& $Bi-GRU_{init}$ \cite{huan2020video}&2020&RNN/GRU& V,T,A &	CMU-MOSI, POM& 	F1-Score,P R,Acc,MAE& \textbf{*} Multimodal contextual information is recorded in real time by attention fusion model \textbf{*} This model improves emotion recognition accuracy in time context\\
			\cellcolor{red!20}& Cimtay et al. \cite{cimtay2020cross}&2020&CNN/InceptionResnetV2& FE, PS &	LUMED-2, DEAP&	WAcc& \textbf{*} Seven emotion states are dealt with from face, EEG, and GSR modalities \textbf{*} Model predict dominant and hidden emotion states with good accuracy\\
			\bottomrule
			
			\multirow{5}{*}{
				\cellcolor{purple!15}}  & $M^{2}DN$ \cite{gao2017event} & 2017 & CNN, HMM & I,T & Brand-Social-Net & R,P, F1-Score & \textbf{*} $M^{2}DN$ is capable of handling weakly labeled microblog data \textbf{*} The Social tracking mechanism extracts the event information from these microblogs\\
			\cellcolor{purple!15} & Huang et al. \cite{huang2018learning} & 2018 & CRBM, CDBN, SVM & V & UCSD, Avenue & AUC,EER, EDR,ACC & \textbf{*} Unsupervised DL model detects anomaly event from the given crowd scenes \textbf{*}Energy, low-level visual and motion map features increase the accuracy of event detection 
			\\
			\cellcolor{purple!15} MMED & Koutras et al. \cite{koutras2018exploring} & 2018 &	CNN, C3D & A,V & COGNIMUSE  & AUC & \textbf{*} C3D-CNN approach detects salient events from the visual stream, while the 2D-CNN approach detects salient events from the audio stream \\
			\cellcolor{purple!15}  & SMDR \cite{yang2019shared} & 2019 & CNN/VGGNet-AlexNet & I, T & MED, SED & NMI, F1-Score & \textbf{*} SMDR model is used to detect real-life events from heterogeneous data \textbf{*} Class-wise residual models in SMDR are designed to detect events with higher accuracy
			\\
			\bottomrule
			
		\end{tabular}
	\end{adjustbox}
\end{table}

\subsubsection{Multimodal Event Detection (MMED):}
\label{EventDetection}

Due to the popularity of media sharing on the internet, users can easily share their events, activities, and ideas anytime. The aim of multimodal event detection (MMED) systems is to find actions and events from multiple modalities like images, videos, audio, text, etc. According to statistics, Million of tweets are posted per day, and similarly, YouTube users post more than 30,000 hours of videos per hour. Hence, in many CV applications, automatic event and action detection mechanisms are required from this large volume of user-generated videos. Finding events and actions from this extensive collection of data is a complex and challenging task. It has various applications in the real-world like disease surveillance, governance, commerce, etc and also helps internet users to understand and captures happenings around the world. Summary of various MMED methods are presented in Table \ref{tab:OthersSummary}. 

Researchers have contributed a lot during the DL era and proposed many methods for event and action detection using multiple modalities. Like,
Y. Gao et al. \cite{gao2017event} proposed a method for event classification via social tracking and deep learning in microblogs. Images and text media are fed to a multi-instance deep network to classify events in this framework.       
S. Huang et al. \cite{huang2018learning} proposed an unsupervised method for the detection of anomaly events from crowded scenes using DL architectures. In this model, visual, motion map and energy features are extracted from video frames. A multimodal fusion network utilized these features for the detection of anomaly events by using the SVM model.
In another research to detect salient events from videos, P. Koutras et al. \cite{koutras2018exploring} employed CNN architecture using audio and video modalities. In this framework, a CNN architecture based on C3D nets is used to detect events from a visual stream, and a 2D-CNN architecture is used to detect events from the audio stream. Experimental results show the improvement of the proposed method performance over the baseline methods for salient event detection.
Z. Yang et al. \cite{yang2019shared} proposed a framework for event detection from multiple data domains like social and news media to detect real-world events. A unified multi-view data representation is built for image and text modalities from social and news domains in this framework. Class wise residual units are formulated to identify the multimedia events.

\section{Experimental results comparisons on benchmark Datasets and Evaluation Metrics:}
\label{reslults}

In this section, we provide a summary of the performance and experimental results of several models that have been reported. The following subsections show the results of all the models presented in section \ref{AppsMMDL}.
\subsection{Multimodal Image Description Results:}

\begin{figure}
	\centering
	\tmpframe{\includegraphics[width=\textwidth,height=80mm]{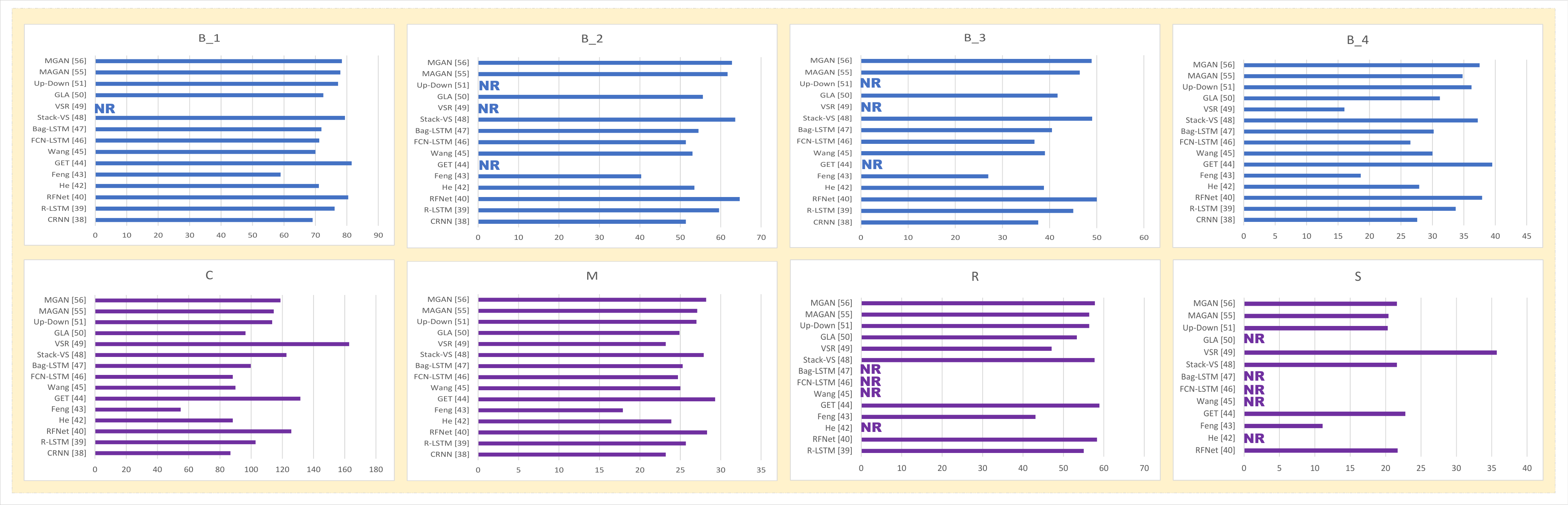}}
	\caption{Experimental results of MMID models on MACOCO dataset, where BLEU (B), CIDEr (C), METEOR (M), ROUGE (R), SPICE (S), and No Result (NR)}
	\label{fig:MMID-Chart results}
\end{figure}

Multimodal image description models are categorized into three parts, i.e., encoder-decoder-based image description, semantic concept-based image description, and attention-based image description. We can discuss the performance of these models on the MS-COCO widely used dataset. Comparative analysis of the MMID model's experimental results on benchmark evaluation metrics is shown in Table \ref{tab:MMID results}.
Visual feature extractor plays an important role in the evaluation of the model's performance. Like for CRNN \cite{wu2017cascade} model, experimental results for BLEU, CIDEr, and METEOR metrics are improved using InceptionV3 visual extractor instead of VGG16. GET \cite{ji2021improving} model achieves the best results on the MS-COCO dataset using all the evaluation metrics as compared to all EDID models. This model gets the best experimental results on encoder-decoder-based image description models and achieves the best results on standard evaluation metrics on semantic concept-based image description and attention-based image description models. Inter and intra layer representations are used to merge local and global features, and the global gated adaptive controller fuses the relevant information at the decoder to enhance the model's performance. The experimental results showed the dominance of GET as compared to other models of image captioning. Experimental results on all listed MMID models are compiled and listed separately for each standard evaluation metric as presented in Figure \ref{fig:MMID-Chart results}. Top row shows values (percentage) of image description models on BLEU (B\_1,2,3,4) metric and bottom row shows values (percentage) on CIDEr (C), METEOR (M), ROUGE (R), SPICE (S) metrics. GET \cite{ji2021improving} model results are better than all comparative image description models on B\_1, B\_4, M, R metrics except results of VSR model on C and S metrics.

\begin{table}[]
	\scriptsize
	\caption{Comparative analysis of MMID model results on MSCOCO dataset using benchmark evaluation metrics, Where B=BLEU, C=CIDEr, M=METEOR, R=ROUGE, S=SPICE, "-" indicates no result}
	\label{tab:MMID results}
	\begin{tabular}{||p{5mm}|p{2cm}p{3cm}lllllp{0.3cm}ll|}
		\bottomrule
		\cellcolor{blue!30} & \textbf{\cellcolor{blue!30}Model} & \textbf{\cellcolor{blue!30}Feature Extractor} & \textbf{\cellcolor{blue!30}B\_1} & \textbf{\cellcolor{blue!30}B\_2} & \textbf{\cellcolor{blue!30}B\_3} & \textbf{\cellcolor{blue!30}B\_4} & \textbf{\cellcolor{blue!30}C} & \textbf{\cellcolor{blue!30}M} & \textbf{\cellcolor{blue!30}R} & \textbf{\cellcolor{blue!30}S} \\
		
		\bottomrule
		\multirow{7}{*}{\cellcolor{orange!25}} & \multicolumn{1}{l}{\multirow{2}{*}{CRNN \cite{wu2017cascade}}} & VGG16 & 69.1 & 51.4 & 37.6 & 27.5 & 86.7 & 23.2 & - & - \\
		\cellcolor{orange!25} & \multicolumn{1}{c}{} & InceptionV3 & 69.5 & 51.8 & 38.0 & 27.7 & 89.4 & 23.5 & - & - \\
		\cellcolor{orange!25} & R-LSTM \cite{chen2017reference} & VGG16 & 76.1 & 59.6 & 45.0 & 33.7 & 102.9 & 25.7 & 55.0 & - \\
		\cellcolor{orange!25} \textbf{EDID} & RFNet \cite{jiang2018recurrent} & ResNet, DenseNet, Inception V3\&V4, Inception-ResNet-V2 & 80.4 & 64.7 & 50.0 & 37.9 & 125.7 & 28.3 & 58.3 & 21.7 \\
		\cellcolor{orange!25} & He et al. \cite{he2019image} & VGG16 & 71.1 & 53.5 & 38.8 & 27.9 & 88.2 & 23.9 & - & - \\
		\cellcolor{orange!25} & Feng et al. \cite{feng2019unsupervised}& InceptionV4 & 58.9 & 40.3 & 27.0 & 18.6 & 54.9 & 17.9 & 43.1 & 11.1 \\
		\cellcolor{orange!25} & GET \cite{ji2021improving} & ResNet101 & 81.5 & - & - & 39.5 & 131.6 & 29.3 & 58.9 & 22.8 \\
		\bottomrule
		
		\multirow{5}{*}{\cellcolor{red!15}} & Wang et al. \cite{wang2018novel} & VGG16 & 70.0 & 53.0 & 39.0 & 30.0 & 90.0 & 25.0 & - & - \\
		\cellcolor{red!15} & FCN-LSTM \cite{zhang2018high}& VGG16 & 71.2 & 51.4 & 36.8 & 26.5 & 88.2 & 24.7 & - & - \\
		\cellcolor{red!15}\textbf{SCID} & Bag-LSTM \cite{cao2019image} & VG16 & 71.9 & 54.5 & 40.5 & 30.2 & 99.8 & 25.3 & - & - \\
		\cellcolor{red!15} & Stack-VS \cite{cheng2020stack} & Faster-RCNN & 79.4 & 63.6 & 49.0 & 37.2 & 122.6 & 27.9 & 57.7 & 21.6 \\
		\cellcolor{red!15} & VSR \cite{chen2021human} & ResNet101 & - & - & - & 16.0 & 162.8 & 23.2 & 47.1 & 35.7 \\
		\bottomrule
		
		\multirow{4}{*}{\cellcolor{blue!15}} & GLA \cite{li2017gla}& Faster-RCNN, VGG16 & 72.5 & 55.6 & 41.7 & 31.2 & 96.4 & 24.9 & 53.3 & - \\
		\cellcolor{blue!15} & Up-Down \cite{anderson2018bottom}& Faster-RCNN, ResNet101 & 77.2 & - & - & 36.2 & 113.5 & 27.0 & 56.4 & 20.3 \\
		\cellcolor{blue!15} \textbf{AID}& MAGAN \cite{wei2020multi} & ResNet & 77.9 & 61.7 & 46.4 & 34.8 & 114.5 & 27.1 & 56.4 & 20.4 \\
		\cellcolor{blue!15}& MGAN \cite{jiang2021multi}& Faster-RCNN & 78.4 & 62.8 & 48.9 & 37.5 & 118.8 & 28.2 & 57.8 & 21.6 \\
		\bottomrule
	\end{tabular}
\end{table}

\subsection{Multimodal Video Description Results:}
\begin{figure}
	\centering
	\tmpframe{\includegraphics[width=\textwidth,height=80mm]{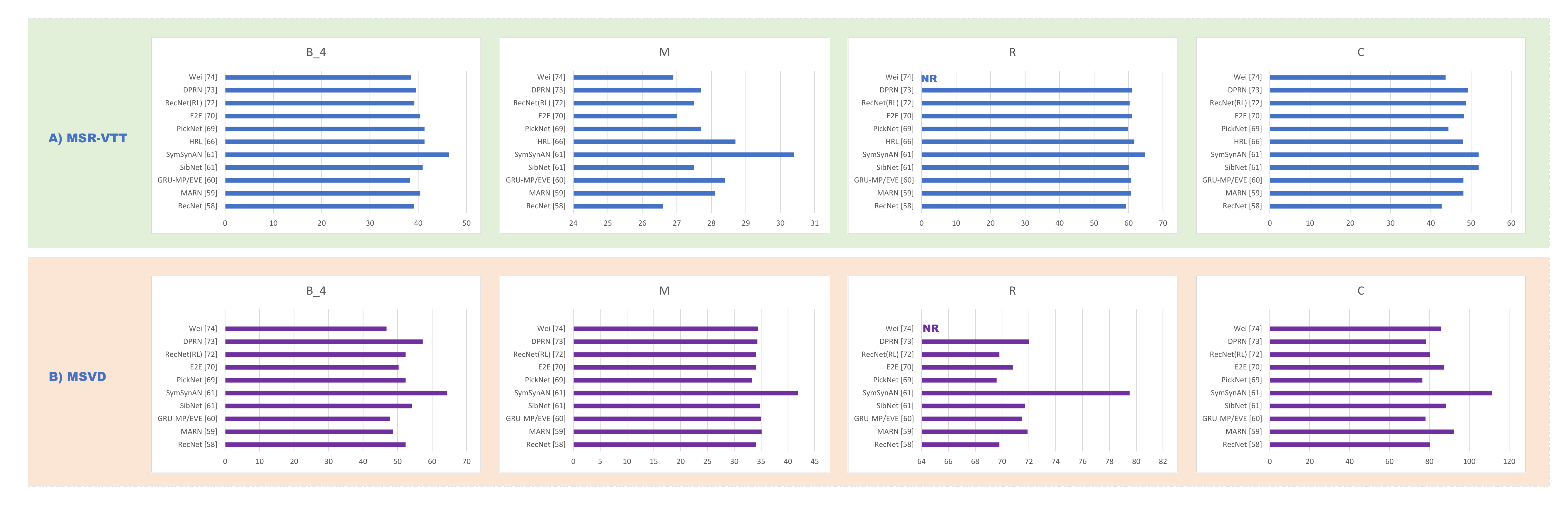}}
	\caption{Experimental results of Video Description models on MSR-VTT and MSVD dataset, where BLEU (B), METEOR (M), ROUGE (R), CIDEr (C), and No Result (NR)}
	\label{fig:vidDescription chart results}
\end{figure}

In this article, multimodal video description models are categorized into CNN-to-RNN encoder-decoder group, RNN-to-RNN encoder-decoder group, and deep reinforcement learning group of applications. Most of these model's performances are evaluated on MSVD and MSR-VTT datasets using BLEU (B), CIDEr (C), METEOR (M), and ROUGE (R) evaluation metrics as shown in Table \ref{tab: VideoResults}. SemSynAN  \cite{perez2021improving} a CNN-to-RNN encoder-decoder based model, achieved the best experimental results on B, C, M, and R metrics as compared to all other video description models. This model outperforms on both MSR-VTT and MSVD datasets. In the SymSynAN model, syntactic representations extract rich visual information from input video frames. Both 2D-CNN and 3D-CNN visual feature extractors are used to extract rich visual features from video and are globally represented. Model guide language model by merging syntactic, visual, and semantic representations to enhance video description accuracy at the decoder stage. Experimental results of all listed MMVD models are compiled and listed separately on common MSR-VTT and MSVD datasets as presented in Figure \ref{fig:vidDescription chart results}. The top row shows values (percentage) of MMVD model's on B, C, M, and R metrics on the MSR-VTT dataset, while the bottom row shows values (percentage) of MMVD model's on B, C, M, and R metrics on MSVD dataset. SemSynAN  \cite{perez2021improving} model clearly shows better results than listed comparative video description models.

\begin{table}[]
	\scriptsize
	\centering
	\caption{Comparative analysis of Multimodal Video description model results on MSR-VTT and MSVD using benchmark evaluation metrics, Where B=BLEU, C=CIDEr, M=METEOR, R=ROUGE, "-" indicates no result}
	\label{tab: VideoResults}
	\begin{tabular}{||p{2.3cm}p{3cm}||p{0.5cm}p{0.5cm}p{0.5cm}p{0.5cm}||p{0.5cm}p{0.5cm}p{0.5cm}p{0.5cm}|}
		\bottomrule
		\multirow{2}{*}{\textbf{\cellcolor{blue!20}}} & \multirow{2}{*}{\cellcolor{blue!20}} & \multicolumn{4}{c}{\textbf{\cellcolor{blue!33} MSR-VTT}} & \multicolumn{4}{c}{\textbf{\cellcolor{blue!42}MSVD}} \\
		\cellcolor{blue!20} \textbf{Model} & \cellcolor{blue!20} \textbf{Feature Extractor} & \textbf{\cellcolor{blue!33}B\_4} & \textbf{\cellcolor{blue!33}M} & \textbf{\cellcolor{blue!33}R} & \textbf{\cellcolor{blue!33}C} & \textbf{\cellcolor{blue!42}B\_4} & \textbf{\cellcolor{blue!42}M} & \textbf{\cellcolor{blue!42}R} & \textbf{\cellcolor{blue!42}C} \\
		\bottomrule
		RecNet \cite{wang2018reconstruction} & InceptionV4 & 39.1 & 26.6 & 59.3 & 42.7 & 52.3 & 34.1 & 69.8 & 80.3 \\
		MARN \cite{pei2019memory}& ResNet101, ResNeXt101 & 40.4 & 28.1 & 60.7 & 48.1 & 48.6 & 35.1 & 71.9 & 92.2 \\
		GRU-MP/EVE \cite{aafaq2019spatio} & Inception-ResNet-V2 & 38.3 & 28.4 & 60.7 & 48.1 & 47.9 & 35.0 & 71.5 & 78.1 \\
		SibNet \cite{liu2020sibnet} & GoogLeNet, Inception & 40.9 & 27.5 & 60.2 & 51.9 & 54.2 & 34.8 & 71.7 & 88.2 \\
		SemSynAN \cite{perez2021improving} & ResNet152, ECO & 46.4 & 30.4 & 64.7 & 51.9 & 64.4 & 41.9 & 79.5 & 111.5 \\
		HRL \cite{wang2018video}& ResNet152 & 41.3 & 28.7 & 61.7 & 48.0 & - & - & - & - \\
		PickNet \cite{chen2018less}& ResNet152 & 41.3 & 27.7 & 59.8 & 44.4 & 52.3 & 33.3 & 69.6 & 76.5 \\
		E2E \cite{li2019end}& Inception-ResNet-V2 & 40.4 & 27.0 & 61.0 & 48.3 & 50.3 & 34.1 & 70.8 & 87.5 \\
		RecNet(RL) \cite{zhang2019reconstruct}& InceptionV4 & 39.2 & 27.5 & 60.3 & 48.7 & 52.3 & 34.1 & 69.8 & 80.3 \\
		DPRN \cite{xu2020deep}& ResNeXt & 39.5 & 27.7 & 61.0 & 49.2 & 57.3 & 34.3 & 72.0 & 78.3 \\
		Wei et al. \cite{wei2020exploiting}& ResNet152 & 38.5 & 26.9 & - & 43.7 & 46.8 & 34.4 & - & 85.7 \\
		\bottomrule
	\end{tabular}
\end{table}

\subsection{Multimodal Visual Question Answering Results:}

VQA is a system where a machine provides the answer in the form of words or short phrases by combining CV and NLP domains. In this review article, MMVQA models are categorized into MMJEM, MMAM, and MMEKM. 
Table \ref{tab:VQA results} shows the performance of VQA models by presenting findings of experimental results conducted on the VQA and VQA family dataset on two test splits termed "test-std and test-dev" to determine the accuracy. VQA is a pretty big dataset comprised of 265,016 pictures and open-ended image-related questions. A strong grasp of language, vision, and common sense knowledge is required to answer these questions. MCAN \cite{yu2019deep} achieves the best overall accuracy on test-dev and test-std as compared to other comparative VQA models. MCAN is a multimodal attention-based network, where the MCA layer uses image features and questions self-attention for the reasoning part. This MCA layer refines the question and image representations to improve the overall accuracy of the model.

\begin{table}[]
	\centering
	\scriptsize
	\caption{Comparative analysis of Multimodal VQA model results on VQA dataset using benchmark evaluation metrics, where "-" indicates no result}
	\label{tab:VQA results}
	\begin{tabular}{||ll||llll|llll|}
		\bottomrule
		\multirow{2}{*}{\textbf{\cellcolor{blue!20}}} & \multirow{2}{*}{\cellcolor{blue!20}} & \multicolumn{4}{c}{\textbf{\cellcolor{blue!20}test-dev}} & \multicolumn{4}{c}{\textbf{\cellcolor{blue!20}test-std}} \\
		\textbf{\cellcolor{blue!20}Model}& \textbf{\cellcolor{blue!20}Feature Extractor} & \textbf{\cellcolor{blue!20}Y/N} & \textbf{\cellcolor{blue!20}Num} & \textbf{\cellcolor{blue!20}Others} & \textbf{\cellcolor{blue!20}All} & \textbf{\cellcolor{blue!20}Y/N} & \textbf{\cellcolor{blue!20}Num} & \textbf{\cellcolor{blue!20}Others} & \textbf{\cellcolor{blue!20}All} \\
		
		\bottomrule
		MUTAN \cite{ben2017mutan}& ResNet152 & 82.88 & 44.54 & 56.50 & 66.01 & - & - & - & 66.38 \\
		MuReL \cite{cadene2019murel}& ResNet152 & 84.77 & 49.81 & 57.85 & 68.03 & - & - & - & 68.41 \\
		\multirow{4}{*}{Wang et al. \cite{wang2017vqa}} & VGG-Open ended & 81.2 & 37.7 & 50.5 & 61.7 & 81.3 & 36.7 & 50.9 & 61.9 \\
		& VGG-Multiple Choice & 81.3 & 39.9 & 60.5 & 66.8 & 81.4 & 39.0 & 60.8 & 67.0 \\
		& ResNet-Open ended & 81.5 & 38.4 & 53.0 & 63.1 & 81.4 & 38.2 & 53.2 & 63.3 \\
		& ResNet-Multiple Choice & 81.5 & 40.0 & 62.2 & 67.7 & 81.4 & 39.8 & 62.3 & 67.8 \\
		MFB \cite{yu2017multi}& ResNet152 & 84.1 & 39.1 & 58.4 & 66.9 & 84.2 & 38.1 & 57.8 & 66.6 \\
		Up-Down \cite{anderson2018bottom}& Faster RCNN, ResNet101 & 81.2 & 44.21 & 56.05 & 65.32 & - & - & - & 65.67 \\
		MCAN \cite{yu2019deep}& ResNet101 & 86.82 & 53.26 & 60.72 & 70.63 & - & - & - & 70.90 \\
		ReGAT \cite{li2019relation}& ResNet101 & 86.08 & 54.42 & 60.33 & 70.27 & - & - & - & 70.58 \\
		Guo et al. \cite{guo2020re}& Faster RCNN & 87.0 & 53.06 & 60.19 & 70.43 & 86.97 & 52.62 & 60.72 & 70.72 \\
		\bottomrule
	\end{tabular}
\end{table}


\subsection{Multimodal Speech synthesis results:}

Comparative analysis of MM speech synthesis model's results on U.S. English speech database is presented in Table \ref{tab:DLTTS results}. For evaluation of speech synthesis, experimental results Mean Opinion Score (MOS) and Mel-Cepstral Distortion (MCD) are the most commonly used subjective and objective evaluation metrics. Subjective assessment approaches are often better suited for assessing speech synthesis models, but they demand substantial resources and confront problems in terms of findings validity, reliability, and reproduction of results. In contrast, the quantitative assessment of the Text To Speech model is performed by the objective evaluation technique. The discrepancies between produced and real speech data are commonly utilized to assess the model. These assessment measures can only represent the model's data parsing capacity to a limited level and cannot indicate the produced speech's quality. The MOS is the most widely employed subjective assessment technique, which evaluates naturalness by inviting participants to rate the synthetic speech. MOS higher scores indicate better speech quality. MCD is the most commonly used objective assessment method to determine the difference between actual and reproduced samples. 
TACOTRON2 \cite{shen2018natural} DLTTS model performs better as compared to other speech synthesis models. Parallel TACOTRON \cite{elias2020parallel} is also the closest model in terms of experimental results, but its relevancy is much more than TACOTRON2.

\section{Discussion and Future Direction:}
\label{Disc&future}
Issues and limitations of various methods and applications mentioned in section \ref{AppsMMDL} are highlighted below. Their’s possible future research directions are also proposed separately. 

\subsection{Multimodal Image Description:}

In recent years, the image description domain improves a lot, but there is still some space available for improvement in some areas. Some methods did not detect prominent attributes, objects and, to some extent does not generate related or multiple captions. Here, some issues and their possible future directions are mentioned below to make some advancement in the image description.

\begin{itemize}
	\item The produced captions' accuracy depends mostly on grammatically correct captions, which depends on a powerful language generation model. Image captioning approaches strongly depend on the quality and quantity of the training data. Therefore, another key challenge in the future will be to create large-scale image description datasets with accurate and detailed text descriptions in an appropriate manner.
	\item Exiting models demonstrate their output on datasets comprising a collection of images from the same domain. Therefore, implement and test image description models on open-source datasets will be another significant research perspective in the future.
	\item Too much work has been done on supervised learning approaches, and training requires a vast volume of labeled data. Consequently, reinforcement and unsupervised learning approaches have less amount of work done and have a lot of space for improvement in the future.
	\item More advanced attention-based image captioning mechanism or captioning based on regions/multi-region is also a research direction in the future.
\end{itemize}

\begin{table}[]
	\scriptsize
	\caption{Comparative analysis of MM Speech synthesis results on U.S. English speech database using Mean Opinion Score (MOS).}
	\label{tab:DLTTS results}
	\begin{tabular}{||lll|}
		\bottomrule
		\textbf{\cellcolor{blue!25}Model} & \textbf{\cellcolor{blue!25}Architecture} & \textbf{\cellcolor{blue!25}MOS} \\
		\bottomrule
		TACOTRON\_Griffin-Lim \cite{wang2017tacotron} & \multirow{2}{*}{Bidirectional GRU/RNN} & 2.57 $\pm$ 0.28 \\
		TACOTRON\_WaveNet &   & 4.17 $\pm$ 0.18 \\
		DeepVoice \cite{arik2017deep} & GRU/RNN & 2.05 $\pm$ 0.24 \\
		DeepVoice2 \cite{gibiansky2017deep} & GRU/RNN & 2.96 $\pm$ 0.38 \\
		DeepVoice3\_Griffin-Lim \cite{ping2017deep} & \multirow{3}{*}{Griffin-Lim, WaveNet Vocoder} & 3.62 $\pm$ 0.31 \\
		DeepVoice3\_WORLD &  & 3.63 $\pm$ 0.27 \\
		DeepVoice3\_WaveNet &  &   3.78 $\pm$ 0.30 \\
		Parallel WaveNet \cite{oord2018parallel} & CNN & 4.21 $\pm$ 0.081 \\
		TACOTRON2  \cite{shen2018natural} & SIFT, BiLSTM & 4.526 $\pm$ 0.066 \\
		Parallel TACOTRON\_Transformer  & \multirow{2}{*}{GLU, VAE, LSTM} & 4.36 $\pm$ 0.04 \\
		Parallel TACOTRON\_LConv \cite{elias2020parallel} &  & 4.40 $\pm$ 0.04 \\
		\bottomrule
	\end{tabular}
\end{table}

\subsection{Multimodal Video Description:}

The performance of the video description process is enhanced with the advancement of DL techniques. Even though the present video description method’s performance is already far underneath the human description process, this accuracy rate is decreasing steadily, and still, there is sufficient space available for enhancement in this area. Here, some issues and their possible future directions are mentioned below to make some advancement in the video description.

\begin{itemize}
	\item It is expected in the future that humans would be capable of interacting with robots as humans can interact with each other. Video dialogue is a promising area to cope with this circumstance similarly to audio dialogue (like, Hello Google, Siri, ECHO, and Alexa). In visual Dialogue, the system needs to be trained to communicate with humans or robots in a conversation/dialogue manner by ignoring conversational statements' correctness and wrongness. 
	\item In CV, most research has focused on descriptions of visual contents without extraction of audio/speech features from the video. Existing approaches extract features from visual frames or clips for a description of the video. Therefore, extraction of these audio features may be more beneficial for the improvement of this description process. Audio features, like the sound of water splash, cat, dog, car, guitar, etc., provide occasional/eventual information when no visual clue is available for their existence in the video.
	\item Existing approaches have performed end-to-end training, resultantly more and more data is utilized to improve the accuracy of the model. But extensive dataset still does not cover the occurrences of real-world events. To improve system performance in the future, learning from data itself is a better choice and achieves its optimum computational power. 
	\item In the video description process, mostly humans describe the visual content based on extensive prior knowledge. They do not all the time rely solely on visual contents; some additional background knowledge is applied by the domain expertise as well. Augmentation of video description approaches with some existing external knowledge would be an attractive and promising technique in this research area. This technique shows some better performance in VQA. Most likely, this would dominate some accuracy improvement in this domain as well.
	\item Video description can be used in combination with machine translation for auto video subtitling. This combination is very beneficial for entertainment and other areas as well. So this combination needs to be focused on making the video subtitling process easier and cost-effective in the future. 
	
\end{itemize}

\subsection{Multimodal Visual Question Answering:} 

In recent years, the VQA task has gained tremendous attention and accelerated the development process by the contribution of various authors in this specific domain. VQA is especially enticing because it comprises a complete AI task that considers free-form, open-ended answers to questions by extracting features from an image/video and the question asked. The accuracy of VQA research still goes beyond answering the visual-based questions as compared to human equivalent accuracy. Some issues and their possible future directions for the VQA task are mentioned below.

\begin{itemize}
	\item So far, the image/video feature extraction part is almost fixed to a model like, most of the time ImageNet Challenge model is used. In this model, image features  are extracted by the split of frames into uniform boxes that works well for object detection, but VQA needs features by tracking all objects using semantic segmentation. Therefore, some more visual feature extraction mechanisms for VQA tasks need to be explored in the future. 
	\item Goal-oriented datasets for VQA research need to be designed in feature to support real-time applications like instructing users to play the game, helping visually impaired peoples, support for data extraction from colossal data pool and robot interaction, etc. So far, VizWiz \cite{gurari2018vizwiz} and VQA-MED \cite{abacha2019vqa}are two publicly available goal-oriented datasets for VQA research. Therefore, in the future, more goal-oriented datasets need to be built for mentioned above applications. 
	\item Most baseline VQA methods have been tested using traditional accuracy measure that is adequate for the multiple-choice format. In the future, the assessment or evaluation method for open-ended frameworks for VQA tasks is examined to improve these models' accuracy. 
\end{itemize}

\subsection{Multimodal Speech synthesis:}
The DLTTS models use distributed representations and complete context information to substitute the clustering phase of the decision tree in HMM models. Therefore, to produce a better quality of speech synthesis process compared to the traditional method, several hidden layers are used to map context features to high dimensional acoustic features by DLTTS methods. More hidden layers inevitably raise the number of system parameters to achieve better performance. As a result, space and time complexity for system training is also increased. Therefore for network training, a large amount of computational resources and corpora is required. Besides these achievements, there is still room available for DLTTS models in terms of quality improvement in intelligibility and speech's naturalness. Therefore, some issues and their feature research directions are discussed below.

\begin{itemize}
	\item DLTTS approaches usually require a huge amount of high-quality (text, speech) pairs for training, which is time-consuming and an expensive process. Therefore in the future, data efficiency for E2E DLTTS  models training is improved by the publicly availability of unpaired text and speech recordings on a large scale.
	\item Little progress is made in front-end text analysis to extract valuable context or linguistic features to minimize the gap between the text-speech synthesis process. Therefore it is a good direction in the future to utilize specific context or linguistic information for E2E DLTTS systems.
	\item Parallelization will be an essential aspect for DLTTS systems to improve system efficiency because most DNN architecture needs many calculations. Some frameworks proposed in recent years also use parallel networks for training or inference and achieve some good results, but there is still room available to achieve optimum results in the future.  
	\item As for DLTTS application concerns, the use of speech synthesis for other real-world applications like voice conversion or translation, cross-lingual speech conversion, audio-video speech synthesis, etc., are good future research directions. 
\end{itemize}

\subsection{Multimodal Emotion Recognition:}

Analysis of automatic emotions requires some advanced modeling and recognition techniques along with AI systems. For more advanced emotion recognition systems future research based on AI based automated systems constitutes more scientific progress in this area. Some future directions of listed issues of multimodal emotion recognition are mentioned below.

\begin{itemize}
	\item Existing baseline approaches are successful, but further experience, knowledge, and tools regarding the analysis and measurement of automatic non-invasive emotions are required. 
	\item Humans use more modalities to recognize emotions and are compatible with signal processing; machines are expected to behave similarly. But the performance of automated systems is limited with a restricted set of data. Therefore, to overcome this limitation, new multimodal recordings with a more representative collection of subjects are needed to be considered in the future.
	\item The preprocessing complexity of physiological signals in emotion detection is a big challenge. In physiological signals, detection of emotion states through electrocardiogram, electromyography, and skin temperature is still emerging. Hence, to determine the potency of these techniques a detailed research can be carried out in the future.
\end{itemize}

\subsection{Multimodal Event Detection:}

Due to the exponential increase in web data, multimodal event detection has attracted significant research attention in recent years. MMED seeks to determine a collection of real-world events in a large set of social media data. Subspace learning is an efficient approach to handle the issue of heterogeneity for the learning of features from multimodal data. On the other hand, E2E learning models are more versatile because they are structured to obtain heterogeneous data correlations directly. MMED from social data is still challenging and needs to be improved in some open issues in the future.

\begin{itemize}
	\item The "curse of dimensionality" issue is raised with the concatenation of various feature vectors of multiple modalities. Some considerable advancement has been made in curse of dimensionality issue, but existing techniques still require some improvement in terms of achieving better learning accuracy for social platforms. Like, GAN or some RNN architectures' extensions are valuable to improve feature learning accuracy of multimodal social data.
	\item Textual data also coincide with audio and video media. Therefore, MMED is improved if this textual information is considered jointly with these media.
	\item Current research primarily focuses on the identification of events from a single social platform. A comprehensive understanding of social data events is obtained by synthesizing the information from multiple social platforms. Hence, event detection methods can be implemented simultaneously to examine social data from multiple platforms using a transfer learning strategy.
	
\end{itemize}

\section{Conclusion:}
\label{conclusion}
In this survey, we discussed the recent advancements and trends in MMDL. Various DL methods are categorized into different MMDL application groups and explained thoroughly using multiple media. This article focuses on an up-to-date review of numerous applications using various modalities like image, audio, video, text, body gesture, facial expression, and physiological signals compared to previous related surveys. A novel fine-grained taxonomy of various MMDL methods is proposed. Additionally, a brief discussion on architectures, datasets, and evaluation metrics used in these MMDL methods and applications is provided. A detailed comparative analysis is provided for each group of applications by discussing the model's architectures, media, datasets, evaluation metrics, features. Experimental results comparisons for MMDL applications are provided on benchmark datasets and evaluation metrics. Finally, open research problems are also mentioned separately for each group of applications, and possible research directions for the future are listed in detail. We expected that our proposed taxonomy and research directions would promote future studies in multimodal deep learning and helps in a better understanding of unresolved issues in this particular area.

%

%

\bibliographystyle{ACM-Reference-Format}
\bibliography{ref}

%

\appendix

\end{document}